%% file: emnlp2020.tex
\documentclass[11pt,a4paper]{article}
\usepackage[hyperref]{emnlp2020}
\usepackage{times}
\usepackage{latexsym}

\usepackage{microtype}
\usepackage{amsmath}

\usepackage{tabularx,ragged2e}
\usepackage{graphicx}
\usepackage{amssymb}
\usepackage{xcolor}
\usepackage{caption}
\usepackage{microtype}  
\usepackage{booktabs}
\usepackage{multirow}
\usepackage{arydshln}
\definecolor{Gray}{cmyk}{0,0,0,0.17}
\usepackage{enumitem}
\usepackage{color, colortbl}
\usepackage{makecell}
\usepackage{subfigure}
\usepackage{pbox}
\usepackage{blindtext}
\hypersetup{
    urlcolor=magenta,
}

\aclfinalcopy 

\renewcommand{\thefootnote}{\fnsymbol{footnote}} 
\title{Video2Commonsense: Generating Commonsense Descriptions to Enrich Video Captioning}

\author{
Zhiyuan Fang\thanks{~~Equal Contribution} \quad Tejas Gokhale\footnote[1]{} \quad Pratyay Banerjee \quad Chitta Baral  \quad Yezhou Yang\\
Arizona State University, Tempe AZ\\
{\tt\small \{zy.fang, tgokhale, pbanerj6, chitta, yz.yang\}@asu.edu} \\

\textbf{\normalsize\url{https://asu-apg.github.io/Video2Commonsense}}
}

\date{}

\begin{document}
\maketitle
\renewcommand{\thefootnote}{\arabic{footnote}} 

\begin{abstract}
Captioning is a crucial and challenging task for video understanding.  
In videos that involve active agents such as humans, the agent's actions can bring about myriad changes in the scene. 
Observable changes such as movements, manipulations, and transformations of the objects in the scene, are reflected in conventional video captioning.
Unlike images, actions in videos are also inherently linked to social aspects such as intentions (why the action is taking place), effects (what changes due to the action), and attributes that describe the agent. 
Thus for video understanding, such as when captioning videos or when answering questions about videos, one must have an understanding of these commonsense aspects.
We present the first work on generating \textit{commonsense} captions directly from videos, to describe latent aspects such as intentions, effects, and attributes.
We present a new dataset ``Video-to-Commonsense (V2C)" that contains $\sim9k$ videos of human agents performing various actions, annotated with 3 types of commonsense descriptions.
Additionally we explore the use of open-ended video-based commonsense question answering (V2C-QA) as a way to enrich our captions.
Both the generation task and the QA task can be used to enrich video captions.
\end{abstract}

\section{Introduction}
\begin{figure*}
    \centering
    \includegraphics[width=0.9\linewidth]{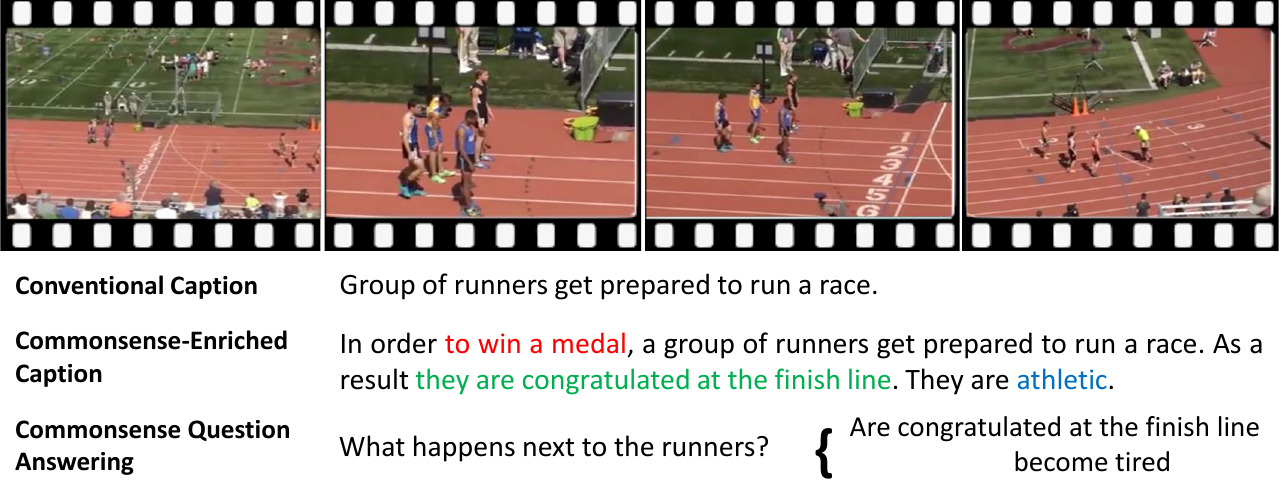}
    \caption{Comparison of conventional video captioning with our commonsense-enriched captioning. Our captions describe intention behind the action ({\color{red}red}), attribute of the agent ({\color{blue}blue}), and effect of the action on the agent ({\color{green}green}).}
    \label{fig:pipeline}
\end{figure*}

When humans watch videos they can typically understand and reason about various aspects of the scene beyond the visible objects and actions.
This involves understanding that some objects are \textit{active agents} that not only perform actions and manipulate objects, but are motivated by intentions, have pre-conditions, and that their actions have an effect on the world and their own mental states.
For instance, in analyzing the video clip in Figure~\ref{fig:pipeline}, humans employ various capabilities such as perception, reasoning, inference, and speculation, to come up with a description for the observable sequence of events, but also reason about latent aspects such as the intention of the group of runners \textit{``to win the medal"}, the effect of being \textit{``congratulated at the finish line"}, and the attribute \textit{``athletic"}.

The above example also illustrates that recognition of objects, actions, and events is often not enough; understanding causal relationships, social interactions, and commonsense aspects behind them provides context and a more semantic interpretation of the video~\cite{gupta2009understanding}.
A model that can provide such detailed interpretations facilitates answering inferential questions, such as \textit{``Will the player get angry later?"}.
However, existing visual understanding systems are unable to perform such tasks that require speculative reasoning.
A critical missing element in complex video understanding is the capability of performing commonsense inference, especially a generative model.
Existing efforts seek to find textual explanations or intentions of human activities as a classification task~\cite{vondrick2016predicting} or a vision-to-text alignment problem~\cite{zhu2015aligning}.

In this paper we propose the \textbf{V}ideo to \textbf{C}ommonsense (V2C) framework to generate visually grounded commonsense descriptions about the underlying event in the video, enriching the factual description provided by a caption.
Under this framework a system is expected to generate captions as well as three types of commonsense descriptions (intention, effect, attribute) directly from an input video.
The V2C model can also be used as a building block for downstream tasks such as video question answering for questions requiring commonsense.
Inspired by~\citep{bosselut2019comet}, our model -- the ``V2C-Transformer" utilizes: (1) a video encoder to extract global representations of the video, (2) a transformer decoder that generates captions and commonsense descriptions, and (3) a cross-modal self-attention module that exploits joint visual-textual embeddings.

We curate the V2C dataset for training and benchmarking models on this task.
We adopt the MSR-VTT video description dataset~\citep{xu2016msr} as a source of videos and captions.
We first utilize the A\textsc{tomic} machine commonsense dataset~\cite{sap2018atomic} to get a list of candidate commonsense texts (intentions, effects, and attributes), and rank these using a BERT-based~\cite{devlin2018bert} model.
Since these candidates are retrieved without using the video and may not be accurate, we instruct humans to watch the videos and select, remove, or rewrite the texts retrieved from A\textsc{tomic}.
The text retrieved by A\textsc{tomic} helps our human annotators to understand the format of desired annotations, and also gives them a list of suggestions.
The human component in our annotation procedure makes our data visually grounded and relevant, linguistically diverse, and natural.

We additionally explore the use of our V2C-Transformer architecture for a open-ended video question answering task, where the questions are about commonsense aspects from the video.
For this, we create a QA addendum of the V2C dataset called V2C-QA.
By asking questions about the latent aspects in the video, our models are able to enrich caption generation with three specific types of commonsense knowledge.

Our contributions are summarized below:
\begin{enumerate}[noitemsep,nosep]
    \item We formulate the ``V2C" task for enriching video captioning by generating descriptions of commonsense aspects.
    \item We curate a video dataset annotated with captions and commonsense descriptions.
    \item We present our V2C-Transformer architecture that generates relevant commonsense descriptions, and serves as a strong baseline.
    \item We pose V2C as a video question answering task and show that it can assist commonsense caption generation.
\end{enumerate}

\section{Video to Commonsense (V2C)}
\paragraph{Problem Formulation:} 
Consider a video $\mathit{V}$ consisting of $N_v$ frames described by sentence $\mathit{S}$. 
Our Video-to-Commonsense (V2C) framework can be used for generating commonsense descriptions $\mathit{C}$ under two settings.
In the first setting (\textbf{V2C-Completion}), we use ground-truth captions to guide commonsense-enriched caption generation.
This task can be viewed as providing supplementary explanations to the caption.
In the second setting (\textbf{V2C-Generation}), we first learn to generate captions from videos, $\mathbf{g}(\mathit{V})$, and then use them to generate commonsense descriptions.
\begin{align}
    \small
    \begin{split}
        \textbf{V2C-Completion}  \quad \mathit{C} &= \mathbf{f}(\mathit{V}, \mathit{S}).\\
        \small
        \textbf{V2C-Generation}  \quad \mathit{C} &= \mathbf{f}(\mathit{V}, \mathbf{g}(\mathit{V})).
    \end{split}
\end{align}

    \subsection{V2C-Transformer}
    \begin{figure*}[t]
        \centering
        \includegraphics[width=.9\textwidth]{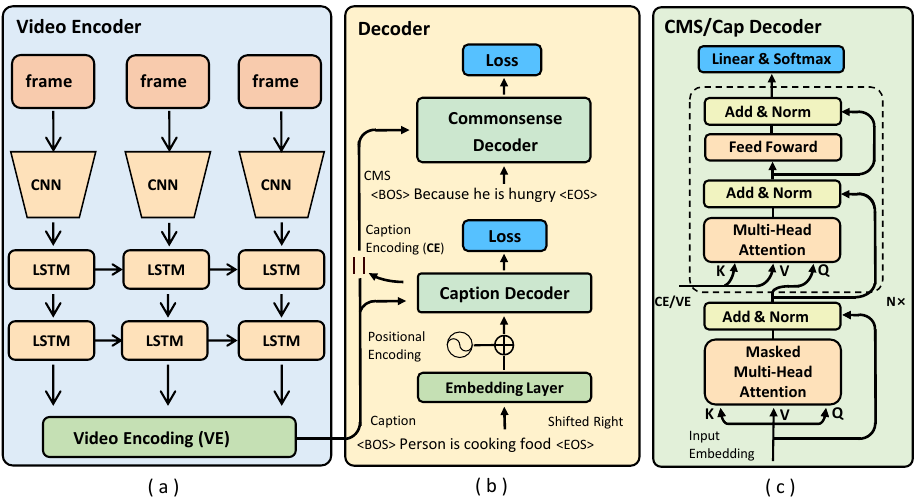}
        \caption{
        The V2C-Transformer model architecture contains:
        \textbf{(a)} Video Encoder designed to take video frames as input and encode them into frame-wise representations, 
        \textbf{(b)} Decoder module consisting of a Caption Decoder and a Commonsense Decoder, and
        \textbf{(c)} Transformer Decoder module containing a stack of $N$ consecutive transformer blocks (shown inside the dashed area).
        }
        \label{fig:architecture}
    \end{figure*}
    
    The proposed Video2Commonsense Transformer is a cross-modal model that generates captions and commonsense-enriched descriptions from videos. 
    Our approach (Figure \ref{fig:architecture}) adopts the ``encoder-decoder'' design: a video encoder that extracts global representations of the input video, and a transformer decoder that produces relevant commonsense knowledge along with captions.
    
    \paragraph{Video Encoder:}
    We obtain per-frame ResNet-152~\cite{he2016deep} features for video $\mathit{V}$ and process them using an LSTM model~\cite{sundermeyer2012lstm}, a standard architecture for modeling long temporal sequences, and use the last hidden states of the LSTM as the video representations.
    We concatenate all previous hidden states from each LSTM module as a final global video encoding $\mathbf{v}$, to provide the model with explicit context using the temporal attention mechanism.
    
    \paragraph{Decoder:}
    The video encoding is used as input to two decoder networks that use a transformer language model~\cite{radford2018improving} to generate a caption and commonsense description, using an inference mechanism similar to~\citet{bosselut2019comet}.
    Our model is a two-stage process that first predicts the current events directly from videos, and then produces the corresponding commonsense captions. 
    During training, the caption decoder $\mathbf{D}_{\textsc{CAP}}$ takes the video encoding ($\mathbf{v}$) and ground truth caption ($\mathbf{s}$) as input to generate caption encoding ($\mathbf{\hat{s}}$), while the commonsense decoder $\mathbf{D}_{\textsc{CMS}}$ uses the concatenation of video and caption encoding to obtain the commonsense description ($\mathbf{c}$), as shown in Figure~\ref{fig:pipeline} (b).
    This arrangement enables the attention module in commonsense decoder to attend to both the video and caption context. 
    \begin{equation}
    \small
        \mathbf{\hat{s}} = \mathbf{D}_{\textsc{CAP}}(\mathbf{v},  \mathbf{s}),  \quad 
        \mathbf{c} = \mathbf{D}_{\textsc{CMS}}(\mathbf{v},  \mathbf{\hat{s}}).
    \end{equation}

    \noindent\textbf{Transformer Decoder}
    is composed of a stack of transformer blocks (dashed area in (c) Figure~\ref{fig:architecture}), whose main component is a self-attention architecture.
    It takes as input the summation of word embedding and the positional encoding offset by 1 position through masked multi-head attention, which prevents the future words been seen. 
    In our model, we deploy two stacked decoder architectures for both caption decoding and commonsense knowledge decoding.
    The Transformer Block consists of consecutive linear transformation: a multi-head attention module (denoted as $\mathcal{H}_{\textsc{M-Att}}$), a two-layer feed forward network ($\mathcal{H}_{\textsc{FFN}}$), a layer normalization operation, and a residual connection.
        
    \paragraph{Multi-head Attention module} 
    To enable our transformer decoder to generate commonsense descriptions by using both the visual and textual content, we modify the multi-head attention module (which acts as the basic unit in recent transformer based language generation models~\cite{radford2018improving,radford2019language}) as a cross-modal module.
    $\mathcal{H}_{\textsc{M-Att}}$ takes the input of the embedding of key (K), value (V) and query (Q).
    The key and value in transformer block are the video encoding (caption decoder) or concatenation of video/caption encoding (commonsense decoder), while the query is the output from the previous transformer block. 
    In the masked multi-head attention module, K, V and Q are the identical vectors of input embedding.
    For a self-attention block with $h$ heads,
        \begin{equation}
        \small
            \mathcal{H}_{\textsc{M-Att}}(\textsc{K}, \textsc{V}, \textsc{Q}) = \mathcal{H}_{\textsc{FFN}}([x_1,\dots, x_h]),
        \end{equation}
        where $x_i$ is computed by scaled dot-product attention operation, for head-index $i$, key-dimension $d_k$n, and transformation parameters $\textsc{w}_i$. 
        \begin{equation}
        \small
        \begin{split}
            \textbf{for } \mathbf{D}_{\textsc{CAP}}, &\quad {x_i} = \textsc{Softmax}(\frac{\textsc{w}^\textsc{q}_i \textsc{Q}\cdot \textsc{w}^\textsc{k}_i \textsc{K}^\prime}{\sqrt{d_k}})\textsc{w}^\textsc{v}_i \textsc{V}, \\
            \textbf{for } \mathbf{D}_{\textsc{CMS}}, &\quad {x_i} = \textsc{Softmax}(\frac{\textsc{w}^\textsc{q}_i [\mathbf{v},  \mathbf{s}]\cdot \textsc{w}^\textsc{k}_i [\mathbf{v},  \mathbf{s}]^\prime}{\sqrt{d_k}})\textsc{w}^\textsc{v}_i \textsc{V}.
        \end{split}
        \nonumber
        \end{equation}


\section{The V2C Dataset}
\begin{figure}[t]
    \centering
    \includegraphics[width=\linewidth]{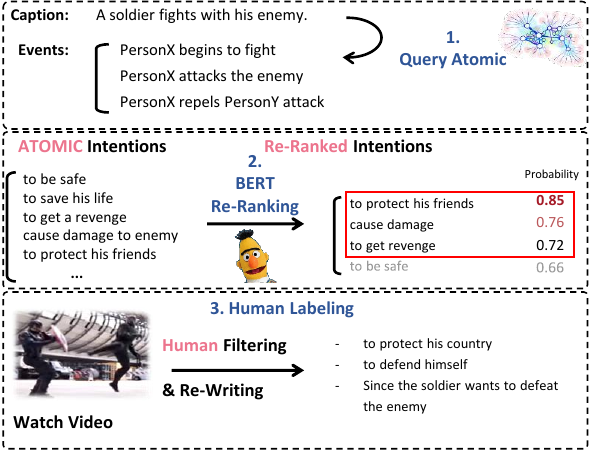}
    \caption{The overall three-step pipeline (retrieval from ATOMIC, BERT re-ranking, and human labeling) to construct our V2C dataset.}
    \label{fig:v2cdataset}
\end{figure}

For the V2C task we need video clips annotated with commonsense descriptions about the agents in the video, as shown in Figure~\ref{fig:pipeline}.
While there are video captioning datasets such as  MSR-VTT~\cite{xu2016msr}, the captions in these datasets describe only the observable objects in the image, but do not describe latent and commonsense aspects.
We are the first to curate such a dataset with annotations describing the intention of agent to perform an action, the effect of the action and the attribute of the agent given the action.

\paragraph{M{SR-VTT}} contains around 10k videos each 10 to 30 seconds long, belonging to 20 categories covering a variety of topics such as sports, music, news, and home videos.
Each video is accompanied by 20 human-annotated textual descriptions on average.
For training and benchmarking the novel V2C task,  we further complement MSR-VTT with event-level commonsense annotations, i.e. event descriptions with intentions, effects and attributes. 
We remove captions and videos that do not have clear human activities.
This is because having such videos leads to an imbalance in the number of captions for each video, thus making it inappropriate to just evaluate caption generation using BLEU scores.

\paragraph{A\textsc{TOMIC}}\cite{sap2018atomic} is an atlas of everyday commonsense knowledge and contains 880k triplets about causes and effects of human activities, organized as \textit{if-then} relations, annotated by crowd-sourced workers.
This data can be categorized based on causal relations, thereby giving us the categories ``cause", ``effect" and ``attribute", e.g., ``\textit{if} X wants to relax, \textit{then} he will play video game."
    
    \subsection{Querying from ATOMIC and Re-ranking}
    \input{tables/atomic_retrieved}
    
    Since inferential knowledge in A\textsc{tomic} only covers human activities, we first retain only those captions in {M{sr-vtt}} that describe human activities.
    We then select three queries from A\textsc{tomic} most similar to the caption, and extract the commonsense descriptions corresponding to these queries.
    In order to select a more reasonable subset of commonsense descriptions, we first train a ranking model.
    We use the BERT~\cite{devlin2018bert} architecture for the ranking model, trained on the ATOMIC dataset for a binary classification task, to predict the relevance of a candidate commonsense description with respect to the event.
    We select the top three relevant intentions, effects, and attributes for each caption.
    This allows us to obtain a preliminary set of 9 commonsense annotations per video directly from the A\textsc{tomic} dataset, relevant to the caption, albeit with noise and annotations that are not relevant to the video.

\input{tables/auto_eval}
    \subsection{Detailed Human Annotation}
    Since we do not use the video to retrieve commonsense descriptions from ATOMIC, we employ human workers to annotate our dataset.
    We recruit two sets of human workers to watch the video, read the caption and select/annotate the relevant commonsense descriptions for each video.
    The first set is Amazon Mechanical Turkers (AMT) who select relevant descriptions.
    The second set is skilled human annotators, screened from a set of university students proficient in English, who are asked to provide annotations in their own words, and remove or edit irrelevant annotations that were provided by ATOMIC and AMT workers.
    This makes our annotations not only grounded in the video, but also more descriptive, linguistically diverse, and of higher quality (see Figure~\ref{fig:v2cdataset}).
    The descriptions from ATOMIC, although not relevant to the video in some cases, give our workers an idea about the format of annotations desired.
    The skilled humans reported that $95\%$ of the captions were relevant, and $65\%$ of the ATOMIC descriptions were useful in understanding the annotation task. Through this procedure, we obtain 6819 videos for training and 2906 videos for testing, a total of 121,651 captions ($\sim$12 captions/video), each caption accompanied with 5 commonsense knowledge annotations (V2C-Raw set). In experiment, we use video captioning technique to conduct the V2C completion task on V2C-Raw set.
    In addition, we instruct human annotators to select and rewrite one raw phrase into complete sentences that complement the captions.
    In total we have 3 complete sentences per video for intention/effect/attribute respectively, and this yields a subset that allows our model to generate complete story-like sentences (V2C-Clean Set).
    Table~\ref{tab:atomic_generations} shows examples from the newly compiled dataset. 
    We conduct rigorous human evaluation to evaluate the quality of our V2C dataset (``Gold Annotations'' in Table~\ref{tab:humanevaluation}).
    Details about the dataset creation process and quality control mechanisms can be found in the Appendix.

\input{tables/human_eval} 
\section{Experiments}
In this section we describe the loss function used for training our model, additional details about video pre-processing, hyper-parameters, and baseline models, and the metrics used for evaluation.

    \paragraph{Loss Function:}
    The decoder parameters $\mathbf{\Theta}$ are trained to maximize the log-likelihood over the training set given by
    $\mathcal{L} = \mathcal{L}_{cap} + \mathcal{L}_{cms}$, where
    \begin{equation}
        \small
        \begin{split}
            \mathcal{L}_{cap}   &= \sum_{t=1}^{N_S}\textrm{log Pr}(\textbf{y}_t| \textbf{y}_{t-1}, \mathbf{v}; \mathbf{\Theta}), and\\
            \mathcal{L}_{cms}   &= \sum_{t=1}^{N_C}\textrm{log}\textrm{ Pr}(\textbf{y}_t|\textbf{y}_{t-1}, [\mathbf{v}, \widetilde{\mathbf{s}}]; \mathbf{\Theta}).
        \end{split}
    \end{equation}
    $\textbf{y}_{t}$ denotes the one-hot vector probability of each word at time $t$, and $N_S, N_C$ denote the length of the caption and commonsense respectively.

    \paragraph{Setting:} 
    In order to obtain video representations, we uniformly sample 40 frames from each video and extract features using feed ResNet~\cite{he2016deep} pre-trained on Imagenet I\textsc{lsvrc}12 dataset~\cite{deng2009imagenet} and get a 2048-d output from the last layer. 
    We use one-hot input (1-of-\textit{N} encoding) of the text input and pass it through an embedding layer to produce a 1028-d hidden vector. 
    We use independent vocabularies for captioning and commonsense generation with sizes 27,603 and 24,010 respectively. Note that, as the generated 

    \paragraph{Hyperparameters:}
    Our decoder is a lightweight transformer decoder consisting of 6 transformer blocks with 8 attention heads each. 
    We use Adam optimizer with 5000 warm-up steps, and learning rate initialized at $1e$-4, and a dropout probability of 0.1 after the residual layer. 
    Our model is trained on a machine with single NVIDIA 1080-Ti GPU.
    
    \paragraph{Baseline Model:} 
    We compare our method with strong video captioning baseline models like, S2VT~\cite{venugopalan2015sequence}, ``Attention-Enc-Dec''~\cite{gao2017video} -- LSTM based models which reach competitive performing on MSR-VTT dataset.
    and ``Dense Captioning''~\cite{zhou2018end}, which is a transformer based video captioning model. As ``Dense Captioning'' is proposed to generate multiple continuous captions for a long untrimmed videos, we modify this by removing the temporal bounding boxes prediction module, and produce two continuous captions (caption + commonsense sentence) together without corresponded starting and ending time.
    All baselines are trained to predict commonsense descriptions from video on the V2C dataset.
    We do not compare with VideoBERT~\cite{sun2019videobert} which is trained on a limited set of cooking videos and hence non-transferable, and requires individual captions for multiple segments of the video.

    \paragraph{Metrics:}
    We report both the performances evaluated by automatic scores and human evaluations following the protocols from~\cite{bosselut2019comet,sap2018atomic}.
    We evaluate our method using BLEU (n=1-4)~\cite{papineni2002bleu}, Meteor~\cite{banerjee2005meteor}, Rouge~\cite{lin2004rouge}, and perplexity score of the generation on its corpus.
    We further conduct human evaluations using AMT workers, who are asked to identity whether the generated commonsense justifiably completes the events (V2C-completion). 
    We follow the setup in~\cite{sap2018atomic} and randomly sample 100 videos from test set and collect 10 generations for each.
    To guarantee the objectiveness of the human evaluations, we hire 5 workers for each sample, yielding \textbf{30k} ratings in total for each model.

    \subsection{Results}
    
    \paragraph{Natural Language Generation Metrics:}
    We show evaluation of the commonsense completion task in Table~\ref{tab:automaticevaluation}.
    Compared to the baseline model, our method exhibits a consistent and overall improvement on almost all metrics.
    Our V2C-Transformer significantly outperforms the LSTM based model in \cite{gao2017video} by 7.7\% at BLEU-4 for the intention prediction.
    Because the V2C-Transformer and the LSTM model share a similar video encoder, our performance improvement could be attributed to the use of self-attention mechanisms in the transformer block in decoding phase. 
    This observation is consistent with the conclusion from~\cite{bosselut2019comet}, and yields further support to the transformer architecture being suited for commonsense inference tasks.
    Moreover, when compared with DenseCap which has a similar transformer architecture and parameters, our model exhibits better evaluation scores, verifying it as a strong baseline model for the V2C task.
    
    \begin{figure*}[t]
        \centering
        \includegraphics[width=\linewidth]{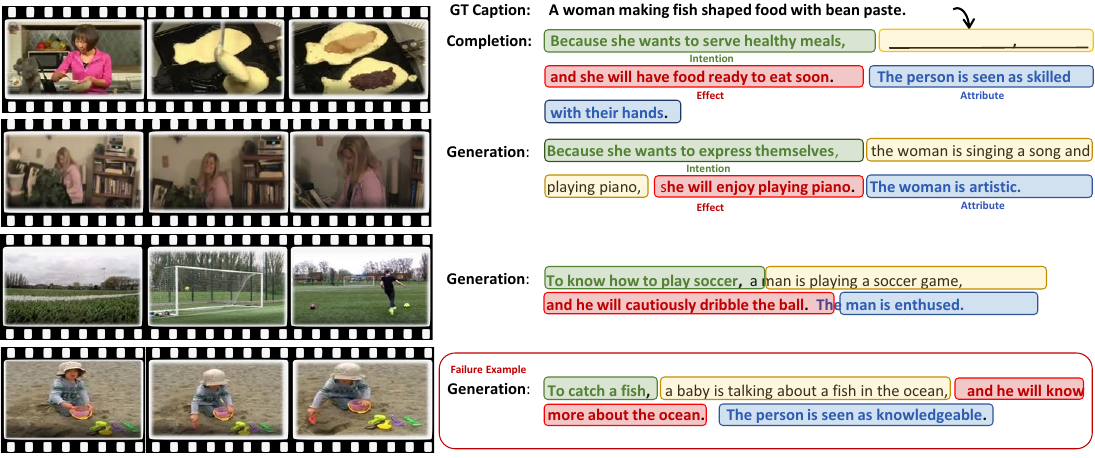}
        \caption{Examples of outputs of our model for the V2C Completion and Generation tasks along with the ground-truth (GT) caption.
        A failure example shown in the bottom red box. }
        \label{fig:qualitative}
    \end{figure*}

    \paragraph{Human Evaluation}
    In Table~\ref{tab:humanevaluation}, {E2C} (Event to Commonsense) is the task of commonsense completion given only textual events~\cite{sap2018atomic,bosselut2019comet} as opposed to V2C which uses both text and video.
    9E\textsc{nc}9D\textsc{ec}~\cite{sap2018atomic} is composed of nine GRU based encoder-decoders as a baseline model for commonsense completion on text, and C\textsc{omet}~\cite{bosselut2019comet} is a large-scale generative pre-trained transformer (GPT) model~\cite{radford2018improving}. 
    We would like to highlight that our transformer model is light-weight with only half of the parameters in GPT without any pre-training. 
    
    We evaluate our model on the tasks of caption generation with human evaluations, and also compare it with the gold annotations. 
    Our gold annotation for ground-truth captions (sourced from the MSR-VTT dataset) points to the fact that a small percentage of captions from MSR-VTT are not relevant to the video, and this is amended by our human workers.
     
    For the  {V2C-Completion} task, our V2C-Transformer model is substantially better (by 7.73\%) than the LSTM-based model from \cite{gao2017video}, and shows consistent lead on each dimension. 
    Thus, when the ground-truth caption is given, our model is able to generate much more relevant commonsense descriptions, thereby consolidating it's ability of commonsense generation.
    
    For the task of  {V2C-Generation}, the difference between human scores for LSTM vs V2C-Transformer is reduced, but our VTC outperforms on average by 2.98\%.
    This may be attributed to the fact that the LSTM-based model is slightly better at generating captions.

    \paragraph{Generating Textual Stories with Commonsense}
    In order to generate story-like textual descriptions that complement the factual captions, we additionally train our model to exploit our diverse complete-sentence annotations.
    Specifically, instead of producing the commonsense knowledge given the videos and captions, we finetune our pre-trained V2C-Transformer model on predicting the human rewritten texts, and generate complete story-like captions.
    Since we do not have enough annotations per sample to compute a fair BLEU score for comparisons, we showcase some sample generated descriptions for qualitative analysis (see Figure~\ref{fig:qualitative}). 
    With that, we observe V2C-Transformer is able to produce complete stories that contain simple, while logically consistent storylines that complement both the visual content and the factual descriptions. 
    We believe that collecting a set of story-like sentences will further enrich our models, and allow us to generate much more contextual, creative, and natural commonsense descriptions from a video.

\section{V2C-QA}
\begin{figure}[t]
    \centering
    \includegraphics[width=\linewidth]{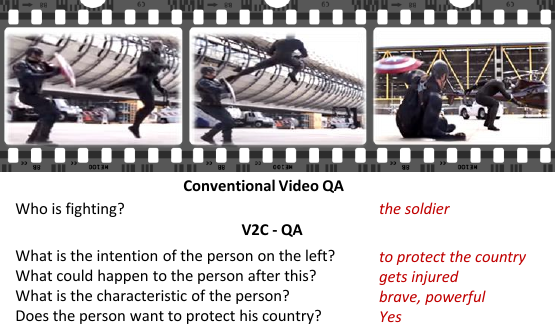}
    \caption{Example questions from V2C-QA compared with conventional video question answering.}
    \label{fig:v2cqa}
\end{figure}
Another way of generating commonsense descriptions about the video is by asking pointed questions.
Consider the example in~\ref{fig:pipeline} where we ask the question \textit{``What happens next to the runners"}, about the \textit{effect} of the action ``prepare" performed by the agents \textit{``group of runners"} observed in the video.
We propose a V2C-QA -- an open-ended commonsense video question-answering task, where we ask questions about the intents, effects and attributes of the agents in the video.

\noindent\textbf{Dataset:}
We use the caption and commonsense annotations in the V2C dataset to create question-answer pairs for each video.
We first extract the action and subject from the caption using SpaCy linguistic features~\cite{honnibal-johnson:2015:EMNLP}.
For each intention, attribute and effect for a video, we use template-based generation to get 7 types of questions -- yielding 21 questions per sample, including negative questions as in~\citet{gokhale2020vqa}.
In total, we have 1,250 training videos and 250 test videos, and a total of 37k questions.
We have a set of 5,555 unique answers for our questions.
Each question can have multiple possible true answers as shown in the example in Figure~\ref{fig:v2cqa}.
The V2C-QA task asks questions that require commonsense reasoning about internal mental states, motivations, and latent aspects of agents in the video as opposed to the conventional video-QA questions about visible objects and actions.

\paragraph{Models:}
We utilize our V2C-Encoder followed by an open-ended answering module.
We jointly predict the type of the question and combine it with the V2C encoding using a feed-forward network.
For textual features, we use embeddings from BERT-base~\cite{devlin2018bert}.
Our models are trained on the open-ended QA task and set-up as a multi-label classification task similar to VQA~\cite{antol2015vqa}, with an answering module design inspired by LXMERT~\cite{tan2019lxmert}.
Our loss function includes the classification loss for answering, the attention loss for question-type, and a label-ranking loss.

\input{tables/v2cqa_results}
\paragraph{Results:}
MSR-VTT QA~\cite{xu2017video} is as a good baseline since it is trained on a conventional videoQA task on the MSR-VTT videos, and only takes video and query as input, unlike recent video understanding models~\cite{lei2018tvqa} that take additional supervision, such as subtitles.
However this model is trained for a multiple-choice QA scheme, so we modify it with our open-ended answering module.
We compare our models when we use our encoder pretrained on the V2C caption generation task, and then finetune it on the V2C-QA task.
We also train models with ground-truth factual captions as input.
Our results are shown in Table~\ref{tab:v2cqa_results}, where we evaluate on prediction of top-k (1,3,5) answers, and report precision and recall.
Our encoder pre-trained on the V2C task outperforms all other models.
Attribute-related questions are easier to answer, while the models struggle the most for questions about intention. Captions help in questions about effects.
The overall text-only baseline shows an insignificant bias between the question and answer-options.

\section{Related Work}
    \noindent\textbf{Video Captioning:}
    Captioning is crucial for understanding visuals; however it is typically limited to describing observable objects and events~\cite{yang2011corpus,thomason2014integrating,gan2017semantic}),
    or for generating paragraphs or multi-sentence captions about the image or video~\cite{krause2016paragraphs,krishna2017dense}.
    However, for detailed video understanding, one needs to obtain descriptions that go beyond observable visual entities and use background knowledge and commonsense to reason about objects and actions.
    Work for inferring motivations of human actions in static images by incorporating commonsense knowledge are reflected in~\citet{pirsiavash2014inferring,vondrick2016predicting}.
    Commonsense caption generation has been approached on abstract scenes and clip-art images in~\citet{vedantamLinICCV15}.
    We present the first generative model for commonsense video captioning.
    
    \noindent\textbf{Video Question Answering:}
    Since caption generation can only describe observable events, recent work seeks to move closer to comprehension, by learning to answer complex questions about videos.
    However, the datasets used for Video QA~\citep{yang2003videoqa,xu2016msr,zhu2017uncovering} focus only on directly evident visual concepts and construct the questions mostly about ``where'' and ``what'' aspects.
    Question answering on movie videos has been explored by~\citet{tapaswi2016movieqa} who collect questions about ``why'' and ``how'' aspects.
    Recently~\citet{lei2018tvqa,zadeh2019social} have propose video-based QA tasks with open-ended high-order questions that need multi-modal understanding, social intelligence modeling, and spatio-temporal reasoning.
    We introduce a novel open-ended video question answering task in this paper, where the questions are about three aspects of commonsense human behavior.

    \noindent\textbf{Visual Reasoning:}    Aspects of visual reasoning have been explored by~\citet{yatskar2016situation} as a situation recognition task on single images, and in Visual Madlibs~\citep{yu2015visual} as a ``fill-in-the-blanks" task for single-image captioning that contains some categories which require reasoning about internal mental states and future events.
    \citet{kim2018textual} provide textual explanations for actions in a self-driving scene.
    \citet{zellers2019recognition} propose a visual question answering task that requires commonsense reasoning to answer a question and to provide a rationale behind the answer.
    Spatial and compositional reasoning is required to answer questions about synthetic images in CLEVR~\cite{johnson2017clevr}. Critical aspects of visual reasoning also include the model's ability to conduct object grounding by natural language descriptions~\citep{rohrbach2016grounding,fang2018weakly,fang2019modularized}.
    Another aspect of visual reasoning is the ability predict a sequence of actions (procedure planning), or to reason about intermediate video frames (walkthrough planning) between two frames, explored in \citet{gokhale2019blocksworld,chang2019procedure}.
    
    \noindent\textbf{Textual Commonsense:}
    Commonsense-based question answering is an area of active research with several datasets and challenges requiring reasoning about conceptual commonsense~\cite{talmor2019commonsenseqa}, physical commonsense~\citep{bisk2020piqa}, social commonsense~\citep{sap2019social}, and abductive commonsense~\citep{bhagavatula2020abductive}.
    On the other hand, challenges such as ProPara~\citep{mishra2018tracking} and bAbI~\citep{weston2015towards} require tracking elements, actions, and effects of actions.
    Commonsense-based text generation has recently been explored via the A\textsc{tomic} dataset~\cite{sap2018atomic}, a corpus of 877k textual descriptions of inferential knowledge organized as \textit{if-then} relations.
    \citet{bosselut2019comet} adopt the A\textsc{tomic} dataset to learn a generative model of commonsense knowledge.
    To the best of our knowledge, ours is the first work on \textit{generating} commonsense descriptions from visual inputs.

\section{Outlook}
A video typically contains one or many objects (sometimes performing actions) in different backgrounds, scenes, or situations.
Some objects may be ``passive" such as trees or buildings, while some objects may be ``active" such as people performing actions like walking, singing, and driving.
This paper is focused on describing such active agents in terms of their intentions, effects of their actions, and attributes that characterize these agents.

We distinguish V2C from the traditional video captioning task.
Video captions describe observable objects, background, and actions, while commonsense descriptions in our task seek to describe the unobservable intentions of the agent (pre-conditions or mental conditions), effects of the action (that happen in the future), and attributes which characterize the agent.
Thus commonsense generation goes \textit{beyond the visible}.
Ours is the first attempt at developing a generative video-based commonsense model.
We anticipate that our framework can be utilized for many applications in video understanding, comprehension, human-robot interaction, and learning commonsense in a multi-modal setting.

\section{Conclusion}
In this paper, we explore a novel and challenging task to generate video descriptions with rich commonsense descriptions that complement the factual captions.
We expand an existing video captioning dataset for the V2C task through automated retrieval from a textual commonsense corpus followed by human labeling, and present a novel V2C-Transformer model to serve as a strong baseline method for the V2C task. 
Our evaluation verifies the effectiveness of our method, while also indicating a scope for further study, enhancement, and extensions in the future. 
Our experiments on using the V2C-Transformer as a component for the V2C-QA task show that the model has transfer learning capabilities that can be applied to other vision-and-language tasks such as question-answering, that require commonsense reasoning.

\section*{Acknowledgements}
{
The authors acknowledge support from the NSF Robust Intelligence Program project \#1816039, the DARPA KAIROS program (LESTAT project), the DARPA SAIL-ON program, and ONR award N00014-20-1-2332.
ZF, TG, YY thank the organizers and the participants of the   \href{https://sites.google.com/view/telluride2019/}{Telluride Neuromorphic Cognition Workshop}, especially the Machine Common Sense (MCS) group.
}
\bibliography{emnlp2020}
\bibliographystyle{acl_natbib}

\clearpage
\appendix
\input{supp}

\end{document}

%% file: tables/atomic_retrieved.tex
\begin{table}[t]
    \begin{center}
      \resizebox{\linewidth}{!}{
        \begin{tabular}{@{}p{0.7cm}c>{\RaggedRight}p{4.3cm}>{\RaggedRight}p{3.2cm}@{}}
        \toprule
        \textbf{Type} & \phantom{a} & \textbf{Video Caption} & \textbf{Commonsense} \\ 
        \toprule
        \multirow{2}{*}{Intention} 
        &  & Two guys are wrestling & to beat the opponent \\
        &  & {A man and woman are singing}  & to express themselves musically \\
         \midrule
         \multirow{2}{*}{Attribute} 
         & & A guy is singing in a crowd  & outgoing\\
         & & Group of riders race on motorcycles. & adventurous  \\
         \midrule
         \multirow{2}{*}{Effect}
         & & A person is making a paper airplane & gets excited to fly it \\
         & & A man and a woman are talking to each other & share ideas and opinions \\
        \bottomrule
        \end{tabular}
        }
    \end{center}
    \caption{Examples of commonsense annotations (intentions, attributes and effects) retrieved from A\textsc{tomic} for captions in MSR-VTT.}
    \label{tab:atomic_generations}
\end{table}

%% file: tables/auto_eval.tex
\begin{table*}[t]
    
    \begin{center}
    \resizebox{\linewidth}{!}{
    \begin{tabular}{@{}p{1.7cm}c  p{5.7cm} c c cccc c c }
        \toprule
        \textbf{Relation} & \hphantom &
        \textbf{Model}  &
        \textbf{CIDER} &             
        \textbf{PPL $\downarrow$} &             
        \textbf{BLEU-1} &
        \textbf{BLEU-2} &
        \textbf{BLEU-3} &
        \textbf{BLEU-4} &
        \textbf{METEOR} &
        \textbf{ROUGE-L} \\
        \toprule
        \multirow{4}{*}{{\normalsize {\textbf{Attribute}}}} 
        && S2VT~\cite{venugopalan2015sequence}  & - & - & 35.9 & - & - & - & - & - \\
        && Attention-Enc-Dec~\cite{gao2017video}  & - & - & 38.3 & - & - & - & - & -  \\
        && Dense Captioner~\cite{zhou2018end}  & - & - & 46.0 & - & - & - & - & -\\
        && Video CMS Transformer & - & - & 47.3 & - & - & - & - & -\\
        \hline
        \multirow{4}{*}{{\normalsize {\textbf{Effect}}}} 
        && S2VT~\cite{venugopalan2015sequence} & 28.3 & 23.6 & 24.9 & 18.6 & 16.2 & 14.3 & 15.4 & 22.1  \\
        && Attention-Enc-Dec~\cite{gao2017video} & 29.5 & 22.0 & 26.5 & 19.4 & 18.8 & 15.1 & 17.5 & 23.9   \\
        && Dense Captioner~\cite{zhou2018end} & 36.9 & 16.0 & 33.7 & 24.8 & 21.0 & 20.2 & 20.0 & 29.9  \\
        && Video CMS Transformer & 37.3 & 15.6 & 34.8 & 25.9 & 22.5 & 20.4 & 20.8 & 30.6  \\
        \hline
        \multirow{4}{*}{{\normalsize {\textbf{Intention}}}} 
        && S2VT~\cite{venugopalan2015sequence} & 51.8 & 17.8 & 48.4 & 39.9 & 34.3 & 26.4 & 23.3 & 44.3\\
        && Attention-Enc-Dec~\cite{gao2017video} & 52.1 & 16.0 & 51.1 & 42.6 & 35.5 & 28.2 & 24.3 & 48.0\\
        && Dense Captioner~\cite{zhou2018end} & 60.3 & 12.0 & 59.3 & 47.0 & 37.3 & 31.5 & 28.0 & 53.1 \\
        && Video CMS Transformer & 62.0 & 11.7 & 60.8 & 48.4 & 39.1 & 34.1 & 28.5 & 54.6  \\
        \hline
    \end{tabular}
    }
    \end{center}
    \caption{ Evaluation of V2C completion task using CIDER, BLEU, Perplexity, Rouge, and Meteor metrics. We use only BLEU-1 to evaluate the attribute generation  since the average length of the ground truth is just less than 2.}

    \label{tab:automaticevaluation}
\end{table*}

%% file: tables/human_eval.tex
    \begin{table*}[t]
        \begin{center}
        \resizebox{0.9\linewidth}{!}{
        \begin{tabular}{c c c c c c c}
            \toprule 
            \textbf{Task} & \multicolumn{1}{c}{\textbf{Model}} & \multicolumn{1}{c}{\textbf{Effect}} & \multicolumn{1}{c}{\textbf{Attribute}} & \multicolumn{1}{c}{\textbf{Intention}} & \multicolumn{1}{c}{\textbf{Average}} & 
            \multicolumn{1}{c}{\textbf{Caption}}\\
            \hline
            \multirow{2}{*}{
                \rotatebox[origin=c]{0}{
                    \shortstack{\textbf{   {E2C-Completion}}\\(Text-Only)}}}
             & 9E\textsc{nc}9D\textsc{ec}~\cite{sap2018atomic} &  44.23 & 52.01 &  49.72 & 49.47 & -\\
             & C\textsc{omet}~\cite{bosselut2019comet} &  54.98 &  56.28 & 66.32 & 59.22 & -\\
             \hline
             \multirow{2}{*}{\rotatebox[origin=c]{0}{\textbf{   {V2C-Completion}}}}
             & Att-Enc-Dec\cite{gao2017video} &  {66.09} & {52.40} & {56.26} & {58.25} & -\\
             & VCT-Completion  &  \textbf{\underline{66.83}} & \textbf{{63.45}} & \textbf{\underline{67.37}} & \textbf{{65.88}} & -\\
             \hline
             \multirow{2}{*}{\rotatebox[origin=c]{0}{\textbf{   {V2C-Generation}}}}
            & Att-Enc-Dec\cite{gao2017video} & {55.93} & \textbf{\underline{74.87}} & {65.54} & {64.78} & \textbf{\underline{74.67}}\\
             & VCT-Generation & \textbf{{62.99}} & 73.54 & \textbf{{66.74}} & \textbf{\underline{67.76}} & 73.17\\
            \hline
            \textbf{   {Gold Annotations}} & V2C Dataset & \textit{75.19}	& \textit{83.03}	& \textit{80.11} & \textit{79.44} & \textit{95.01}\\
            \bottomrule
        \end{tabular}
        }
        \end{center}
           \caption{Human evaluation scores for V2C.
                Captions are an input for the \textbf{   {V2C-Completion}} task, and generated for the \textbf{   {V2C-Generation}} task.
                The best model is given in bold, while the overall best is underlined.
        }
        \label{tab:humanevaluation}
    \end{table*} 

%% file: tables/v2cqa_results.tex
    \begin{table}[t]
        
        \centering
        \resizebox{\linewidth}{!}{
        \begin{tabular}{@{}p{0.45mm}p{3.77cm} p{0.65cm}p{0.65cm} p{0.65cm}p{0.65cm} p{0.65cm}p{0.65cm}}
            \toprule 
            & \multirow{2}{*}{\textbf{Model}} 
            & \multicolumn{2}{c}{\textbf{top-1}}   
            & \multicolumn{2}{c}{\textbf{top-3}}   
            & \multicolumn{2}{c}{\textbf{top-5}} 
            \\
            \cmidrule{3-4} \cmidrule{5-6} \cmidrule{7-8}
            & & p & r & p & r & p & r\\
            
            \toprule 
            \multirow{5}{*}{\rotatebox{90}{\textbf{Intention}}} 
            & MSR-VTT QA        & 9.68 & 2.13 & 7.15 & 4.68 & 6.07 &  6.60\\
            & V2C-T             & 10.34 & 2.31 & 7.69 & 5.03 & 6.37 & 6.87\\
            & V2C-T + Captions  & 10.72 & 2.54 & \textbf{8.08} & 5.47 & 6.39 & 7.20\\
            & Pretrained V2C-T  & 10.77 & \textbf{2.69} & 8.01 & 5.58 & \textbf{6.71} & 7.88\\
            & Pretrained V2C-T~+~Cap. & \textbf{11.04} & 2.68 & 7.96 & \textbf{5.70} & 6.63 &\textbf{7.79}\\
            
            \midrule 
            
            \multirow{5}{*}{\rotatebox{90}{\textbf{Effect}}} & MSR-VTT QA & 19.89 & 5.02 & 8.04 & 5.91 & 5.30 & 6.49\\
            & V2C-T & 20.95 & 5.43 & 8.65 & 6.57 & 5.65 & 7.06\\
            & V2C-T + Captions  & 20.95 & 5.32 & 8.50 & 6.48 & 5.76 & 7.26\\
            & Pretrained V2C-T & 20.95 & 5.32 & 8.63 & 6.55 & 5.82 & 7.49\\
            & Pretrained V2C-T~+~Cap. & \textbf{21.12} & \textbf{5.60} & \textbf{8.70} & \textbf{6.89} & \textbf{5.83} & \textbf{7.68}\\
            \midrule
            
            \multirow{5}{*}{\rotatebox{90}{\textbf{Attribute}}} & MSR-VTT QA & 46.10 & 37.22 & 16.02 & 49.45 & 7.49 & 41.03\\
            & V2C-T & 59.52 & 48.30 & 22.39 & 51.40 & 13.97 & 52.57\\
            & V2C-T + Captions & 59.74 & 48.22 & 23.12 & 52.44 & 14.64 & 54.35\\
            & Pretrained V2C-T & \textbf{60.72} & \textbf{49.00} & \textbf{23.18} & \textbf{52.73} & \textbf{14.98} & \textbf{55.40}\\
            & Pretrained V2C-T +Cap. & 59.57 & 48.24 & 23.10 & 52.54 & 14.94 & 54.91\\
            \midrule 
            & Text-Only Baseline & 12.36 & 11.70 & 13.84 & 12.35 & 14.77 & 14.10 \\
            
            \bottomrule
        \end{tabular}
        }
        \caption{Precision (p) and Recall (r) for V2C-QA for each type of question. 
        }
        \label{tab:v2cqa_results}
    \end{table} 

%% file: supp.tex
\section*{Appendix}
In this appendix, we provide the the following supplementary information:
\begin{itemize}[noitemsep,nosep,leftmargin=*]
    \item Additional details about our dataset creation process, including statistics and analysis for V2C and V2C-QA datasets, 
    \item Examples of commonsense descriptions generated by our V2C-Transformer model, 
    \item Details about our human evaluation interface, protocol, and metrics.
\end{itemize}

\section{V2C Dataset Construction}
\begin{figure*}
    \centering
    \includegraphics[width=\linewidth]{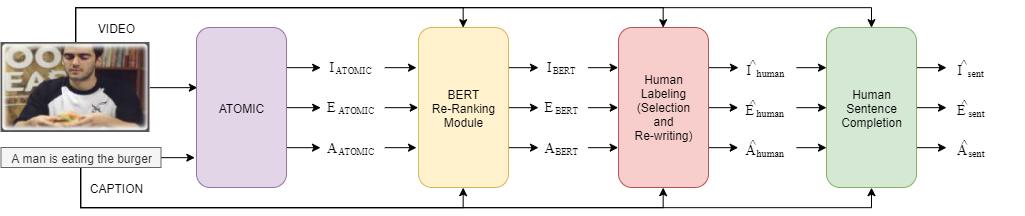}
    \caption{The data creation flow for V2C. We use the retrieved videos and captions from MSR-VTT and use the BERT re-ranking module to obtain a list of top-3 intentions ($I$), effects ($E$), and attributes ($A$).
    These are then further improved by human labeling.
    A subset of annotations is also converted to full sentences by human annotators.}
    \label{fig:flow}
\end{figure*}

Our dataset creation methodology is a three-step procedure as shown in Figure~\ref{fig:flow}.
In the first step, we use the caption to query ATOMIC~\cite{sap2018atomic} and retrieve the top-3 intentions, effects, and attributes.
These are re-ranked by a BERT based model in the second step.
The final step involves humans in the annotation process.
We ask human annotators to select the most relevant descriptions, and to provide additional descriptions in their own words.
The annotators also convert a subset of our dataset into complete sentence descriptions.

    \subsection{Querying from ATOMIC}
    For every video-caption pair in the M\textsc{sr-vtt} dataset, we select 3 most similar events from A\textsc{tomic}.
    These are then used to retrieve textual descriptions of three types -- {\it intentions, effects, attributes} from A\textsc{tomic}.

    \subsection{BERT Ranking Model}
    We implement a Bidirectional Encoder Representations from Transformers (BERT) model~\cite{devlin2018bert} as a ranking model to rank and retrieve \textit{top-3} most plausible commonsense aspects to complement the ground truth caption.
    This is done by treating the ranking task as a binarized next sentence prediction (NSP) task, 
    trained on the A\textsc{tomic}~\cite{sap2018atomic} dataset. 
    When choosing the sentences A and B for each training pair, for 50\% of the training pairs we choose the actual next sentence that follows A, and a random sentence from the A\textsc{tomic} as a negative sentence.
    This setting is consistent with the NSP task in~\cite{devlin2018bert}. 
    We train our model in A\textsc{tomic}, and use it to expand video captions from  M\textsc{sr-vtt}~\cite{xu2016msr}. 
    Our BERT model consists of 12 transformer blocks, 12 attention heads, and 768 hidden dimensions (110M parameters in total). 
    In total, we have 115,312 pairs for training/testing.
    We evaluate our model using accuracy of the prediction in the test set of A\textsc{tomic} which is 30\% of the entire set.
    BERT can achieve 86.21\% accuracy in NSP task on average as shown in Table~\ref{tab:bert}.
    In addition, we also conduct human evaluations to measure the overall quality of the expanded V2C dataset (see ``gold annotations'' in Table. 3, main paper). 
    
    \subsection{Human Labeling}
    
\begin{figure}[t]
    \centering
    \includegraphics[width=\linewidth]{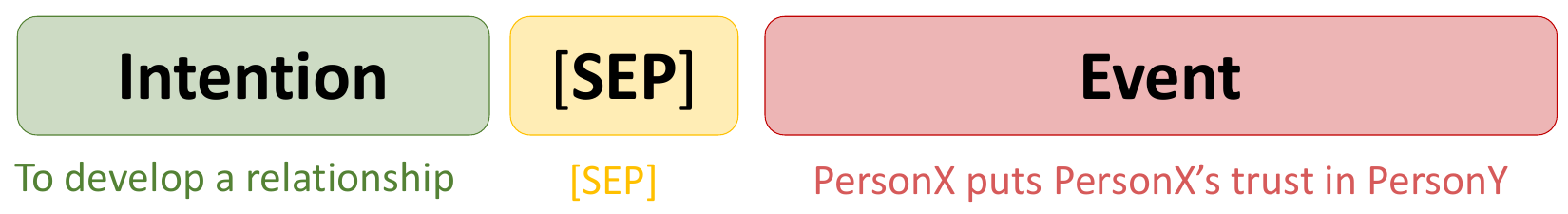}
    \caption{Next sentence prediction task in Bert model. A and B sentences are separated by token {[SEP]}.} 
    \label{fig:bert}
\end{figure}
    
\input{tables/bert_ranking}

\begin{figure}[t]
    \centering
    \includegraphics[width=\linewidth]{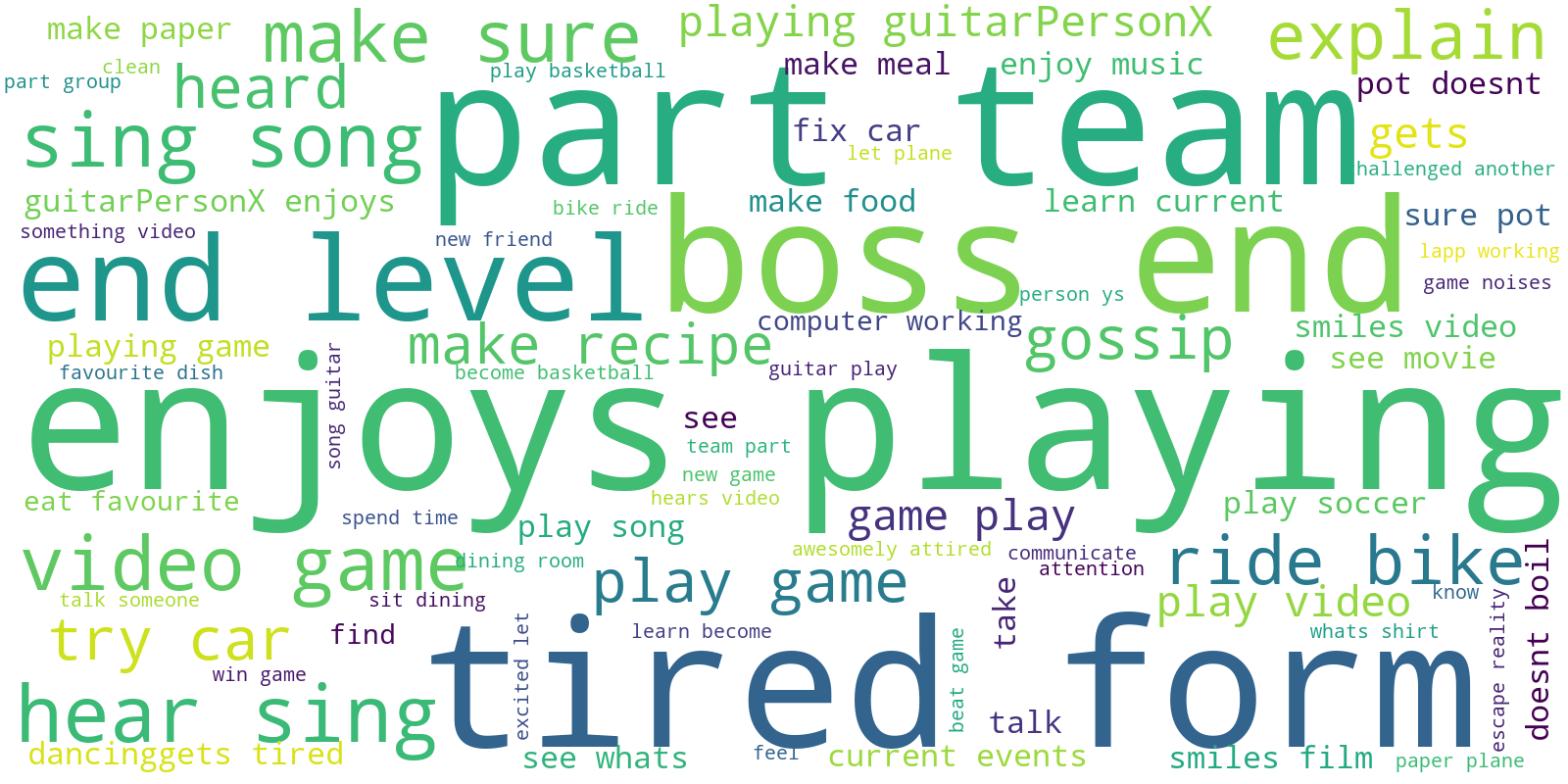}
    \caption{Word cloud figure of the intention commonsense annotations from our V2C dataset.}
    \label{fig:word_cloud}
\end{figure}

\begin{figure}[!h]
    \centering
    \includegraphics[width=\linewidth]{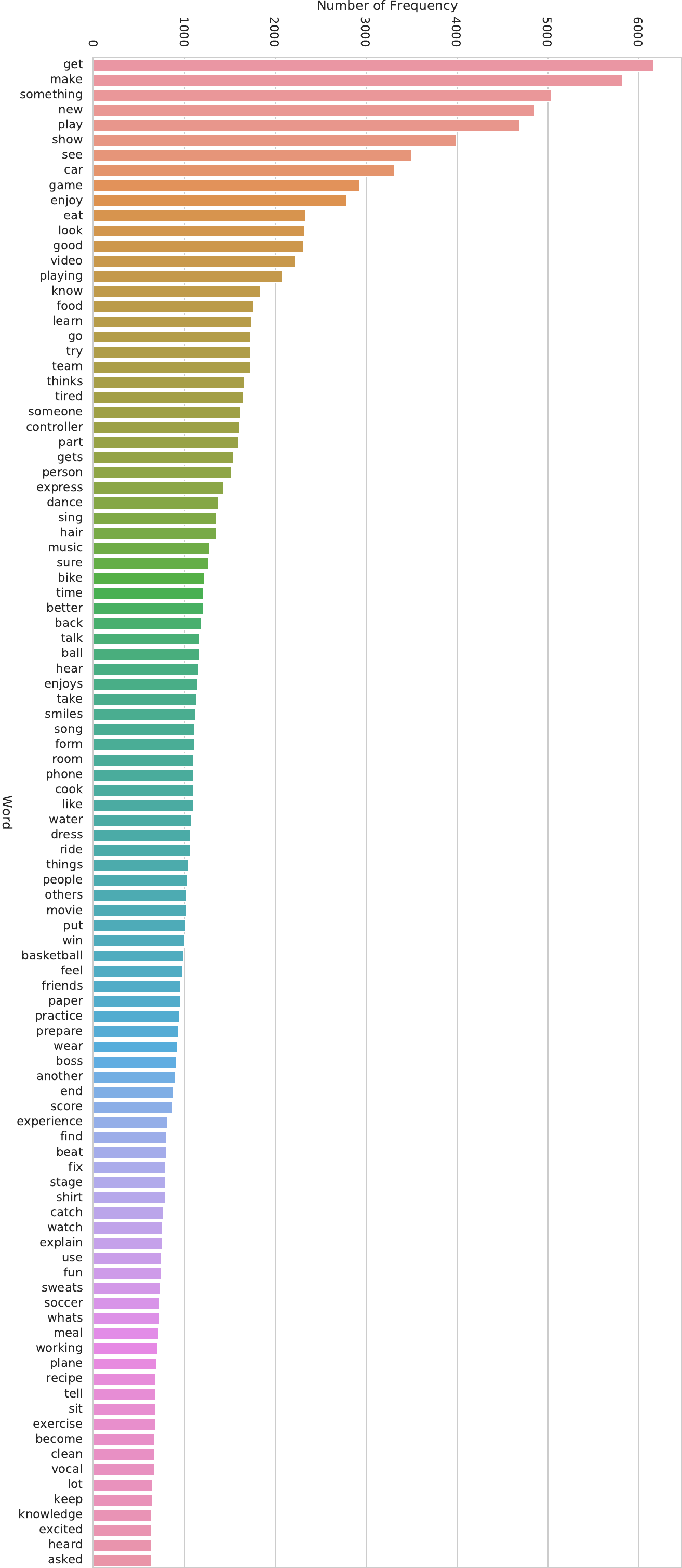}
    \caption{Top-100 most frequent words in our V2C dataset (stop words are ignored).}
    \label{fig:word_freq}
\end{figure}

\begin{figure*}[t]
    \centering
    \includegraphics[width=\linewidth]{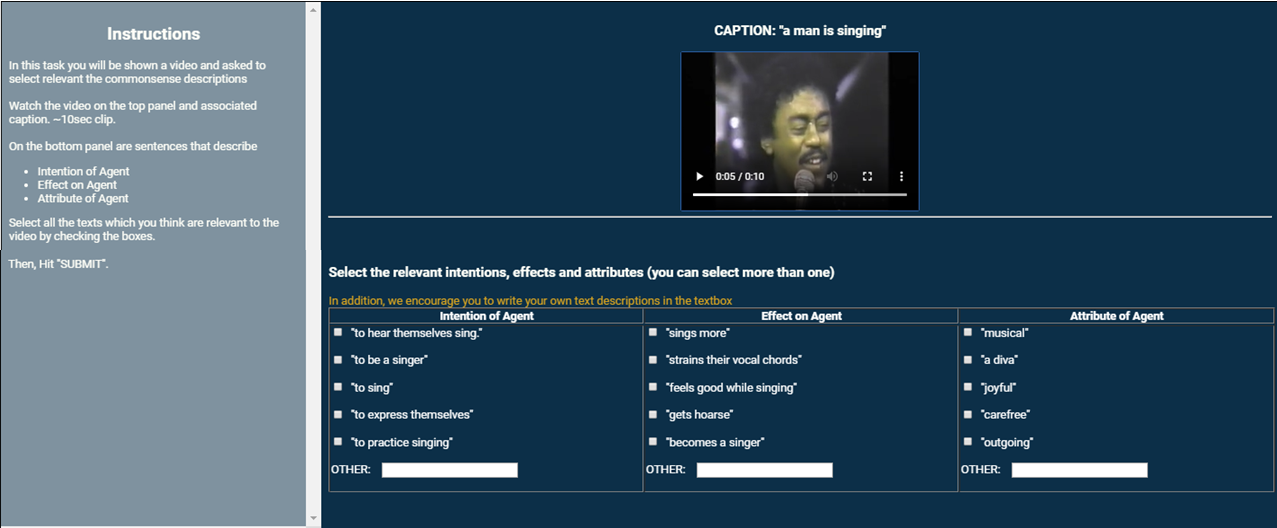}
    \caption{Our human labeling interface. We ask human workers to select relevant commonsense descriptions as well provide additional texts in their own words}
    \label{fig:human_select}
\end{figure*}

\begin{figure*}[t]
    \centering
    \includegraphics[width=0.9\textwidth]{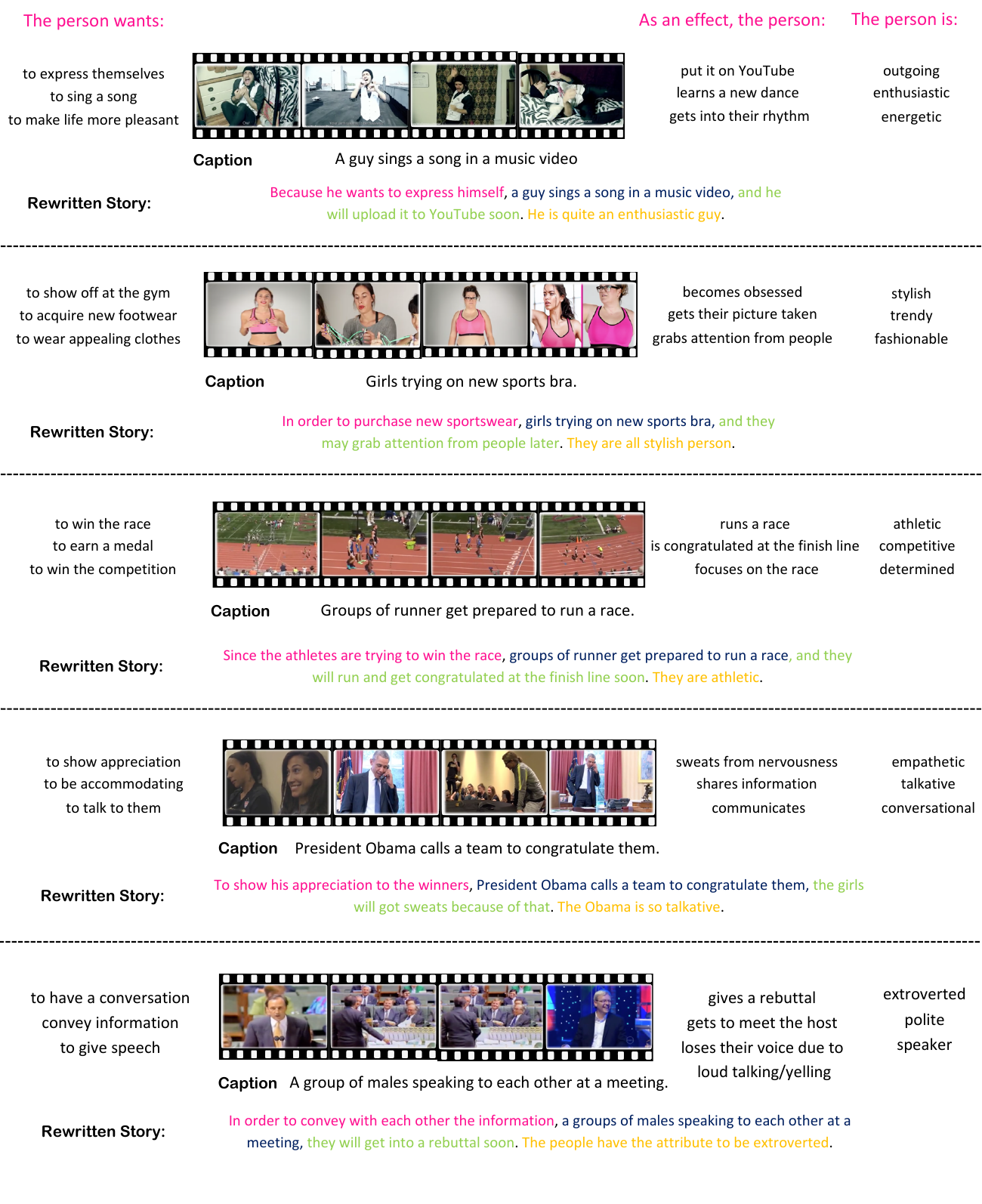}
    \includegraphics[width=0.9\textwidth]{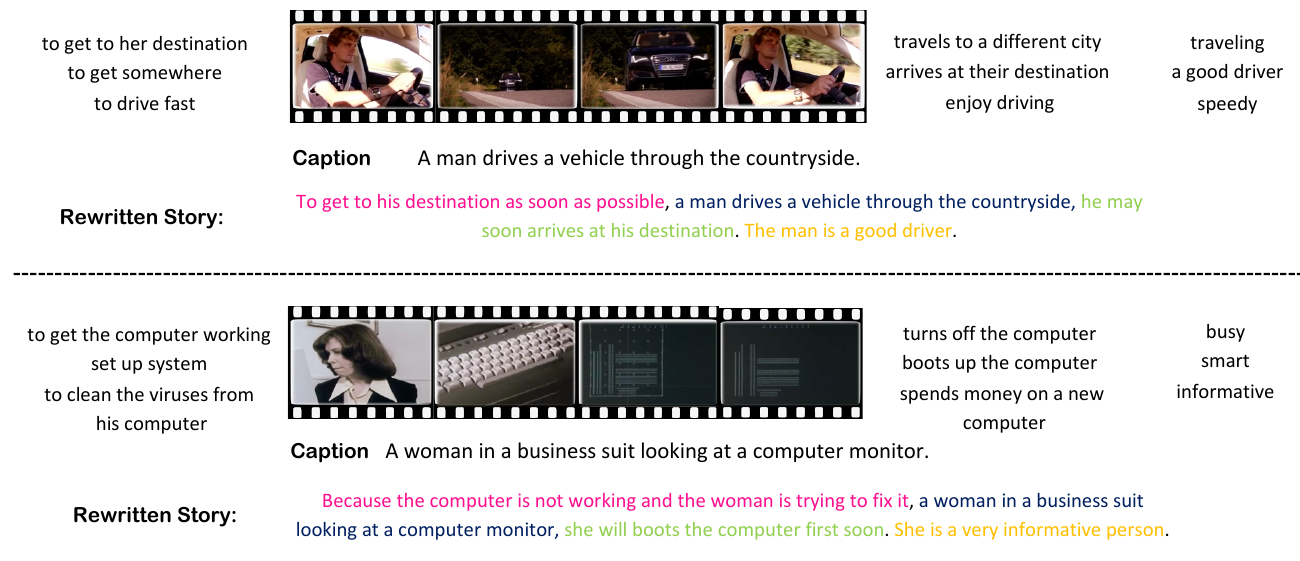}
    \caption{Qualitative examples of our V2C dataset.}
    \label{fig:v2c_example}
\end{figure*}

    With querying from ATOMIC and BERT re-ranking, we obtain commonsense descriptions that are relevant to the caption.
    However, we want to make sure that these descriptions are also relevant to the video.
    Thus we utilize human workers from Amazon Mechanical Turk (AMT) for selecting the most relevant commonsense descriptions.
    Our annotation interface is shown in Figure~\ref{fig:human_select}.
    We ask the annotators to select descriptions that are most relevant to the video and to the caption, and also encourage them to add their own commonsense descriptions.
    This makes our dataset more natural and human-like.
    This also allows us to remove noisy annotations that may be produced due to text-only ATOMIC querying.
We show additional samples from our V2C dataset in Figure.~\ref{fig:v2c_example}, word cloud in 
Figure.~\ref{fig:word_cloud} and word frequency in ~\ref{fig:word_freq}.

\subsection{Benefits of the Three-Step Pipeline} 
Since our videos are annotated with captions, we use the captions to retrieve commonsense descriptions from ATOMIC.
The A\textsc{tomic} dataset has comprehensive annotations for human activities, actions, and events and as such \textbf{covers most of the events in M\textsc{sr-vtt}}. 
Thus using these two datasets together is a natural step for creating our V2C dataset. 

This purely caption-based retrieval unfortunately does not incorporate the latent aspects of the video, but only those from the caption.
Moreover, since the video is not used for retrieving these, the commonsense annotations may be out-of-context.
Thus, we bring in human annotators to watch the video, read the caption, and then use the set of descriptions from ATOMIC to select the relevant once and to discard the irrelevant or out of context descriptions.
The human annotators then provide annotations about intention, effect, and attribute in their own words.
The ATOMIC retrieved descriptions help the human annotators to get an idea about the task and also get a glimpse of the format of the desired annotations.
This significantly reduces the noise in human annotations.

To guarantee and measure the overall quality of our V2C dataset, we have conducted human evaluations on the V2C annotations. Our results shows that 86.29\% of the \big \langle video-caption-commonsense{\big\rangle}  triples are labeled as reasonable samples (see ``Gold Annotations'' in main paper, Table.~3), \textbf{verifying the quality} of our dataset

\input{tables/v2cqa}

\section{V2C-QA Dataset}
For the V2C Question Answering task, we repurpose our V2C dataset and convert it to a question-answering dataset.
We choose a subset of 1500 videos: 1250 for training and 250 for testing, following the same train-test split as MSR-VTT.
We use SpaCy linguistic features~\cite{spacy2} along with the LemmInflect library ({\url{https://github.com/bjascob/LemmInflect}}) and template-based generation to convert the captions, intentions, effects, and attributes from V2C to create questions and ground-truth answers.
Our templates are lingustically diverse, natural, and grammatically sound.
We have 21 types of templates with each template having numerous possibilities for combinations of the slots in the template.
Thus we get 21 types of questions (7 each for intention, effect, and attribute) as shown in Table~\ref{tab:v2cqa_data}.
Since our task is open-ended question-answering, our questions are annotated with \textit{all possible} correct answers for that question.
To get answers for the ``negative" questions as shown in Table ~\ref{tab:v2cqa_data}, we use the adversarial matching strategy similar to~\cite{zellers2019recognition}, by using RoBERTa~\cite{liu2019roberta} similarity.
We will release our V2C-QA question and answer generation code publicly.


\section{Qualitative Generation Results}
We show additional {\textbf{V2C-Completion}} samples by our V2C-Transformer model in Table.~\ref{tab:completion}.
    
\begin{table*}[t]
    \centering
    \resizebox{\linewidth}{!}{
    \begin{tabular}{@{}p{4cm}>{\RaggedRight}c p{5.8cm}>{\RaggedRight}c p{3.6cm}>{\RaggedRight}c  p{1.8cm}>{\RaggedRight}c@{}}
        \toprule
        \textbf{Intention}  &\phantom{a}& \textbf{Caption} &\phantom{a}& \textbf{Effect}  &\phantom{a}& \textbf{Attribute}\\
        \toprule
        to entertain people             &&  a band is performing for a crowd  &&   gets applause &&  acting \\
        to try out PersonY's new car    &&   a man checks out detail on a car &&   gets a speeding ticket &&   helpful \\
        to learn about current events   &&    a complex news host gives an update on rappers. &&    gets informed about current political events &&    talkative \\
        to be in a good mood            &&    a group of people trying to perform an exorcism on a girl &&    gets applause &&    fun \\
        to show his knowledgeable       &&    there is an old man is answering to somebody questions &&    gets another question && sporty \\
        to score a point                &&    a man is shooting a basketball ground &&    gets exercise &&   helpful \\  
        to share their message          &&    a man giving a speech to important people &&    gets applause &&    orator \\  
        to be safe from anything that lurks in the dark &&    a group of people are being chased by crocodiles &&    gets tired from taking pictures &&    scared \\  
        to be informed about the world  &&    a girl is describing about hot news &&    learns about whats happening worldwide && gossipy \\   
        to watch something interesting  &&    a children s television show clip &&    smiles at the screen &&    entertained \\
        to enjoy the evening with the concert band &&     a band composed of older gentlemen are playing blue grass music on a small stage and people are dancing along to the music swing-style &&     gets tired form dancing &&    fun \\  
        to be part of the team && there is a woman playing badminton in a court &&  gets tired after exercise &&    athletic \\  
        to try out person ys new car &&     a boy explaining the features of a car &&     they check car websites online to look at deals &&     helpful \\  
          to escape reality &&     a man explaining a video game &&     takes the video game home &&     gamer \\   
        to cook something &&     there is a man in black cutting the green leaves on the desk &&     gets clean dishes &&     hungry \\
    \bottomrule
    \end{tabular}
    }
    \caption{Illustrative samples generated by our V2C-Transformer model on {\textbf{V2C-completion}} task. }
    \label{tab:completion}
\end{table*}

\input{tables/human_stats}

\begin{figure*}[t]
        \centering
        \includegraphics[width=\textwidth]{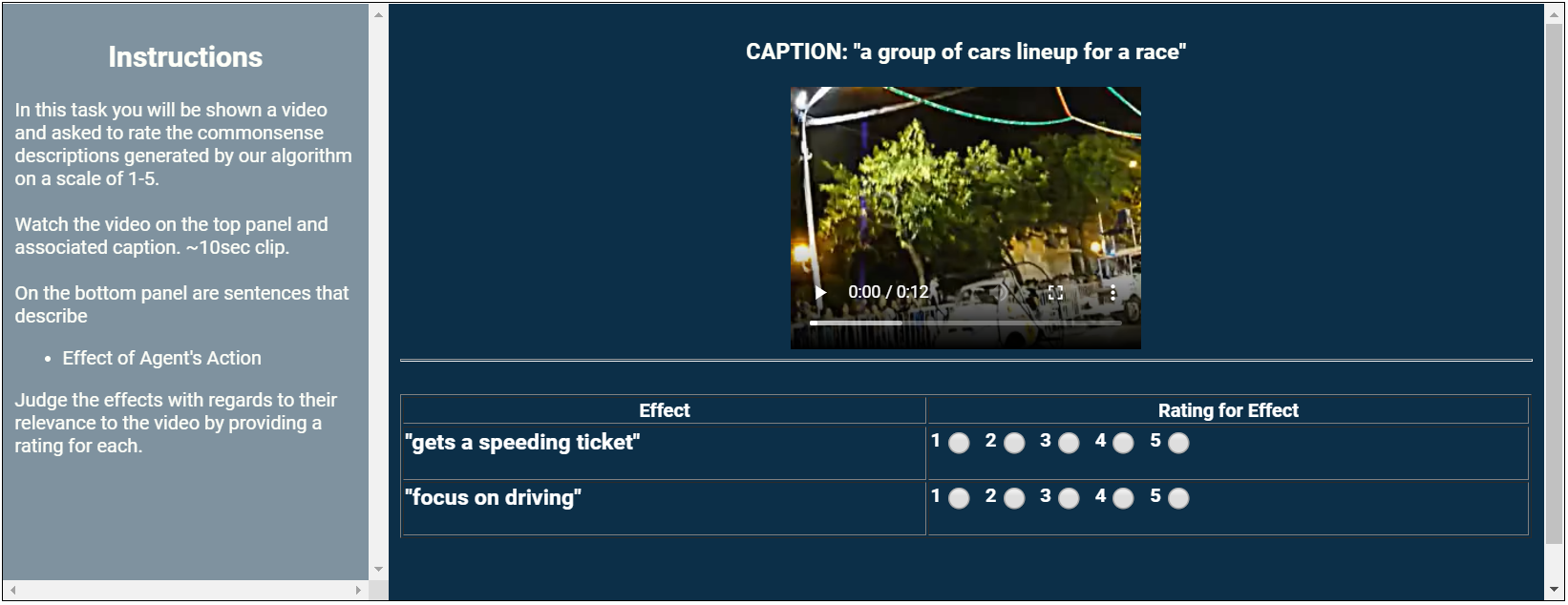}
        \includegraphics[width=\textwidth]{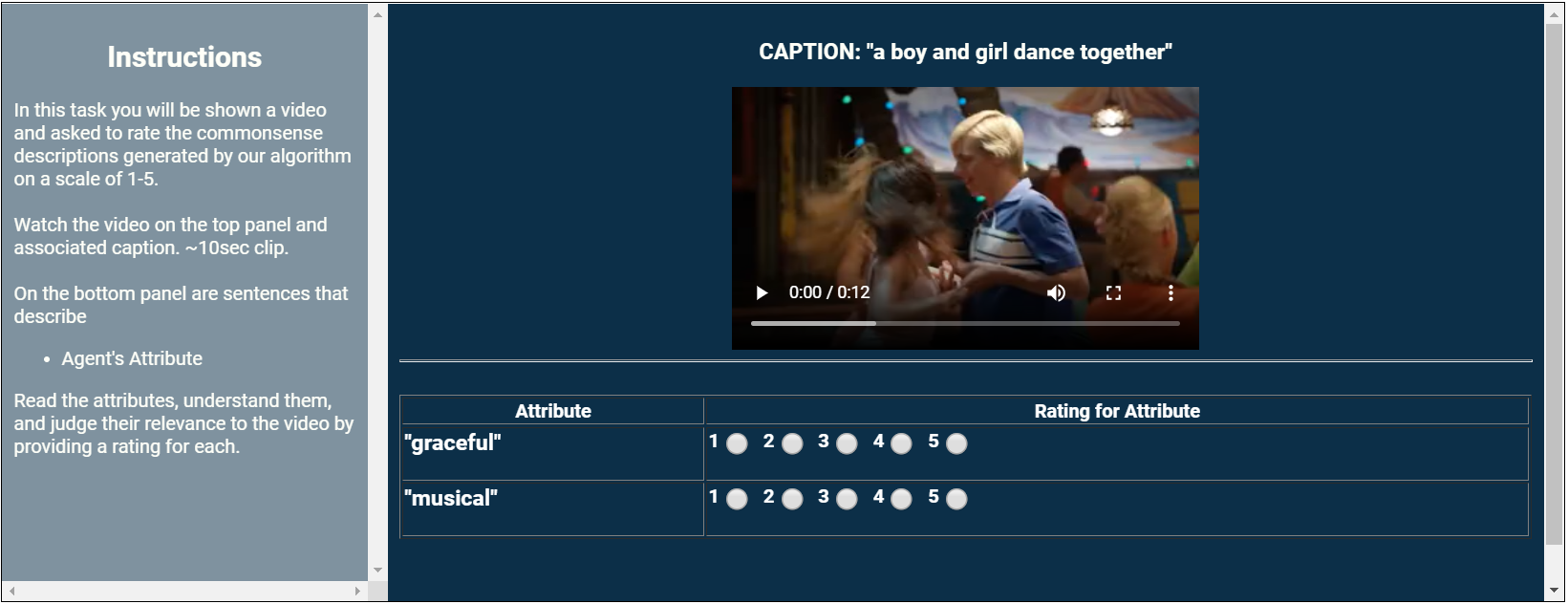}
        \includegraphics[width=\textwidth]{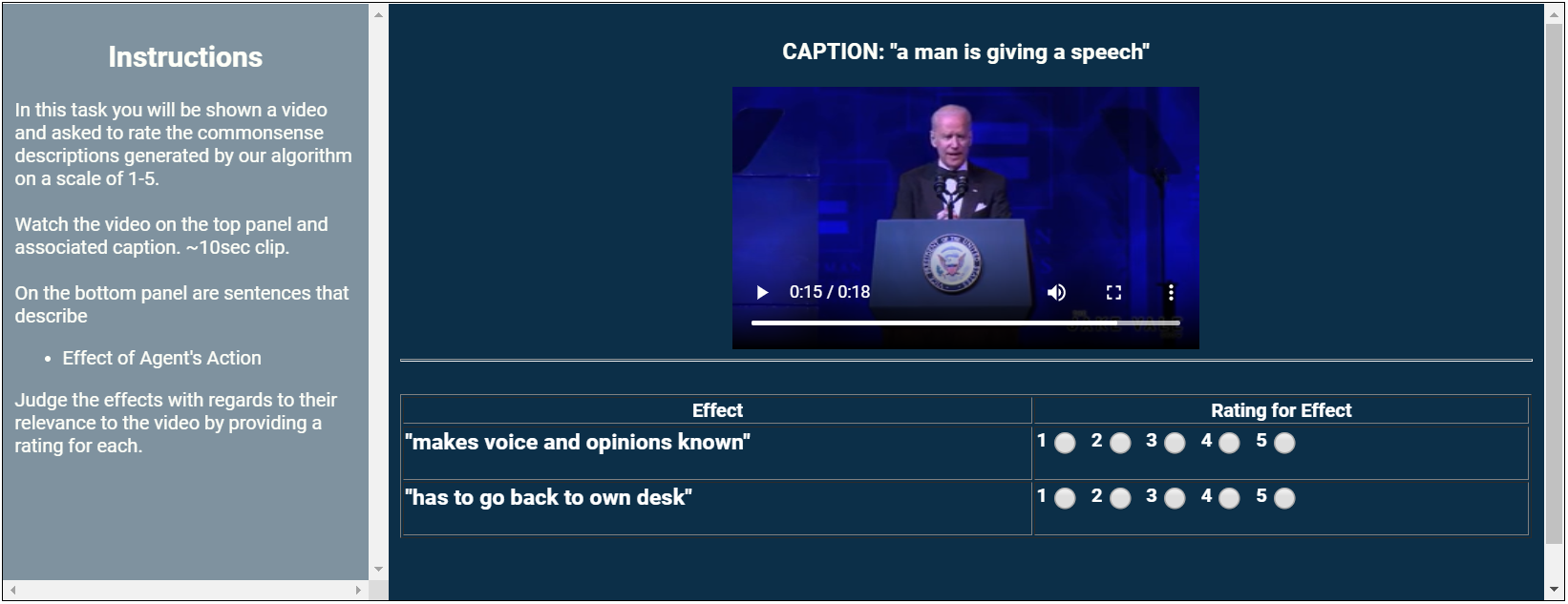}
        \caption{Snapshot of our AMT human evaluation interface for {V2C-completion} task.}
        \label{fig:amt_com}
    \end{figure*}
    
    \begin{figure*}
        \centering
        \includegraphics[width=\textwidth]{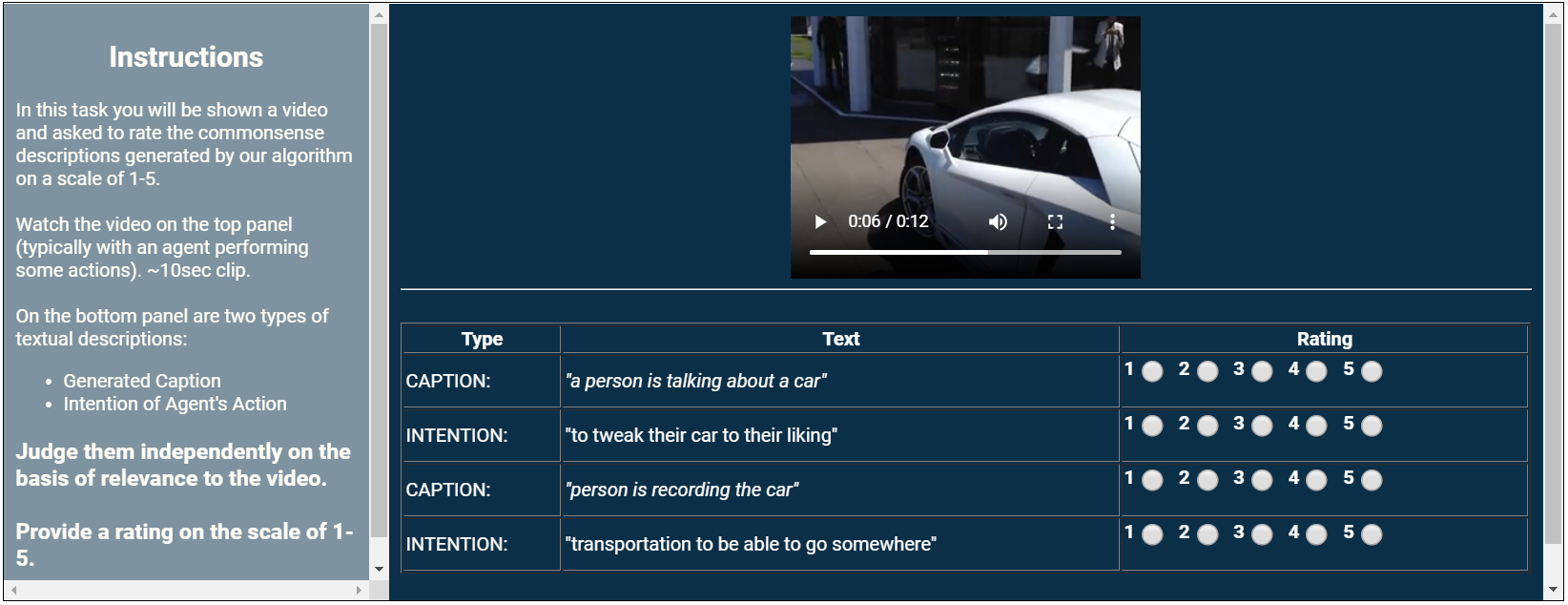}
        \includegraphics[width=\textwidth]{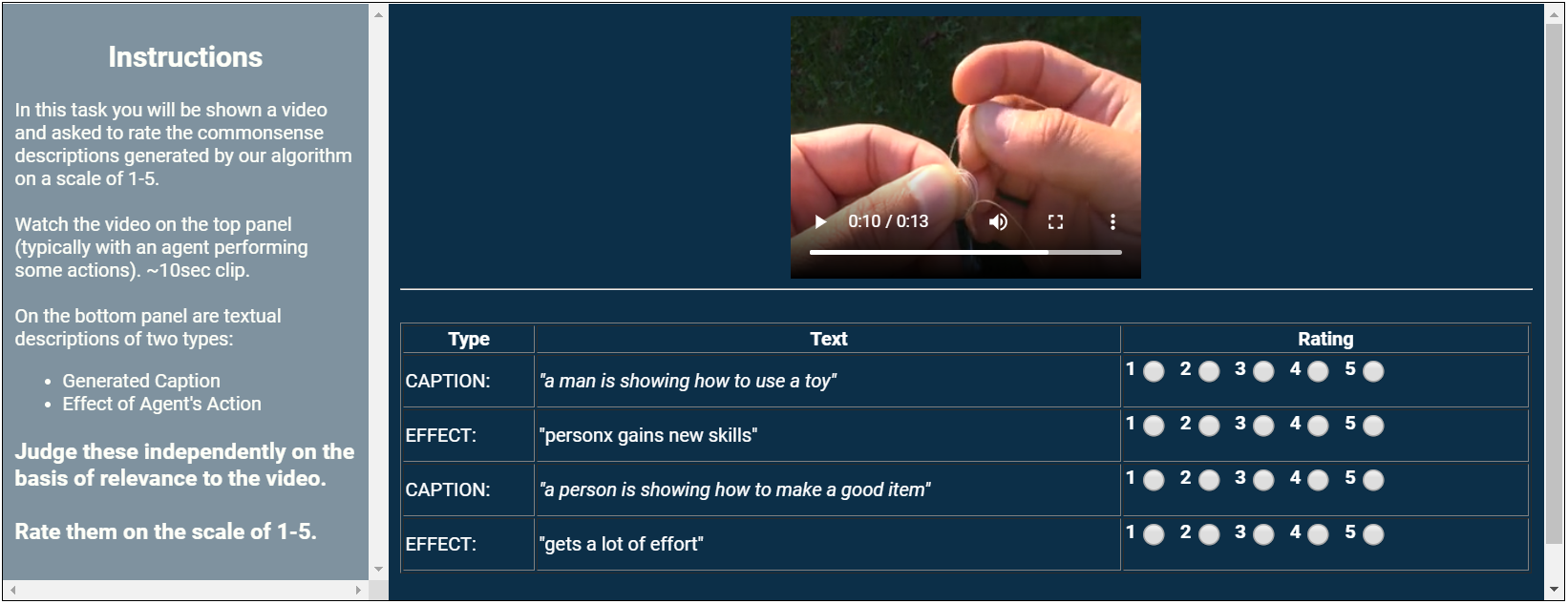}
        \includegraphics[width=\textwidth]{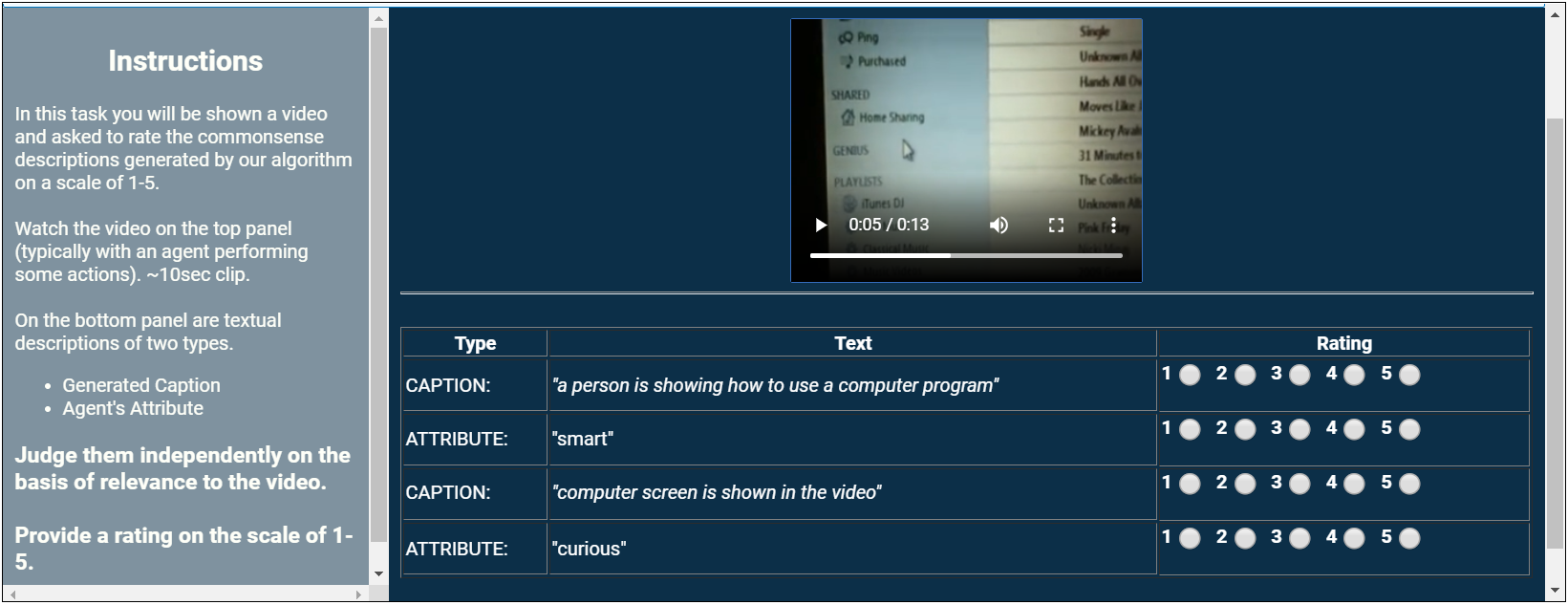}
        \caption{Snapshot of our AMT human evaluation interface for {V2C-generation} task.}
        \label{fig:amt_gen}
    \end{figure*}
    
\section{Human Evaluation}
Human evaluation is one of the important part to verify the performances of our model and the quality of the V2C dataset. 
In this section we describe our setup for human evaluation of the captions and commonsense descriptions in our dataset as well as those generated by our models.

\subsection{Amazon Mechanical Turk Interface}

    We conduct our human evaluations by crowdsourcing ratings from workers on Amazon Mechanical Turk (AMT).
    We do these human evaluations on the same test set used for our automated metrics.
    We show an example of our interface in Figure~\ref{fig:amt_com} and ~\ref{fig:amt_gen} which shows the screenshot of the rating task as seen by the workers.
    The workers are given explicit instructions about this rating task, and depending on the task are asked to rate the commonsense descriptions and the caption. 
    
    For the {V2C-Completion} task, the workers are provided with the video and the ground-truth caption and asked to rate the only the generated commonsense (intention, effect or attribute) on a scale of 1 to 5. The workers are asked to provide this rating on the basis of whether the generated text is relevant to the video, i.e whether the caption/commonsense can plausibly complete the given event. 
    
    For the {V2C-Generation} task, the workers are asked to rate the caption as well as the commonsense texts with respect to the video.
    The workers are also asked to conduct identical tasks for the gold (ground-truth annotations) in our new V2C dataset.

    \subsection{Scheme for Validity}
    Our ratings are measured on a scale of 1 to 5. 
    Annotations which receive a score greater than 3 are considered ``valid", so as to be consistent with the binary ratings used by ~\cite{bosselut2019comet} for their experiments.
    We then compute average validity scores for each commonsense aspect: intention, attribute and effect.

    \subsection{Statistics of Human Evaluations}
        In order to further analyze the human evaluations on our generated outputs, we use three metrics - standard deviation of the ratings, inter-rater agreement score (IRAS) and a smooth version of IRAS.
        Standard Deviation was calculated per sample based on the evaluations provided by multiple workers on each sample. 
        We do so to evaluate how consistent our AMT workers are and how much they deviate or agree with each other.
        We use three different metrics so as to analyze our data and generations through multiple lenses, to be certain that the outputs and annotations are high-quality.

        \begin{figure*}
            \begin{subfigure}
            \centering
            \includegraphics[width=\linewidth]{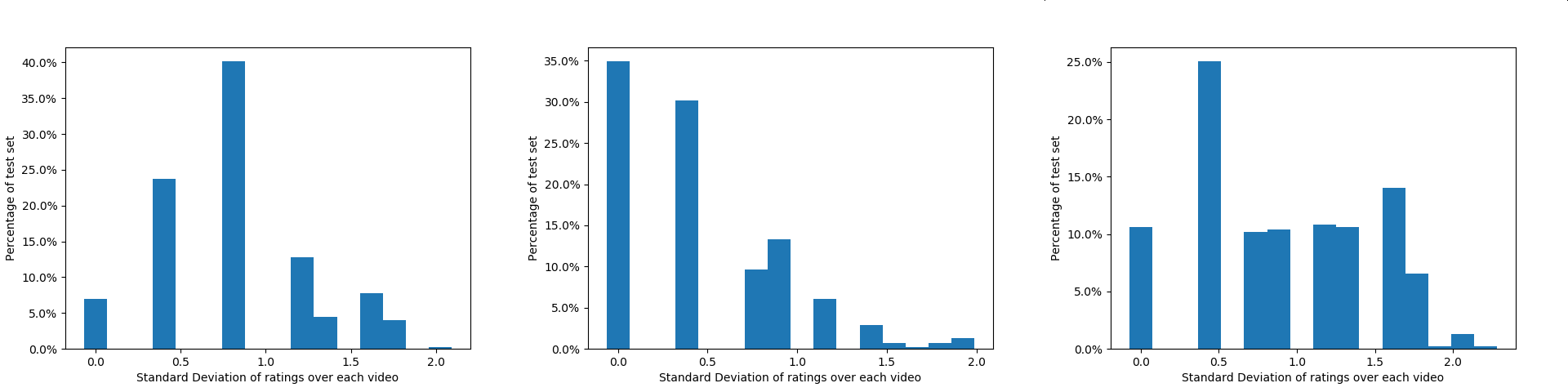}
            \caption{V2C-Completion task using the A\textsc{tt}E\textsc{nc}D\textsc{ec} model.}
            \label{sfig:testa}
            \end{subfigure}\hfill
            \begin{subfigure}
            \centering
            \includegraphics[width=\linewidth]{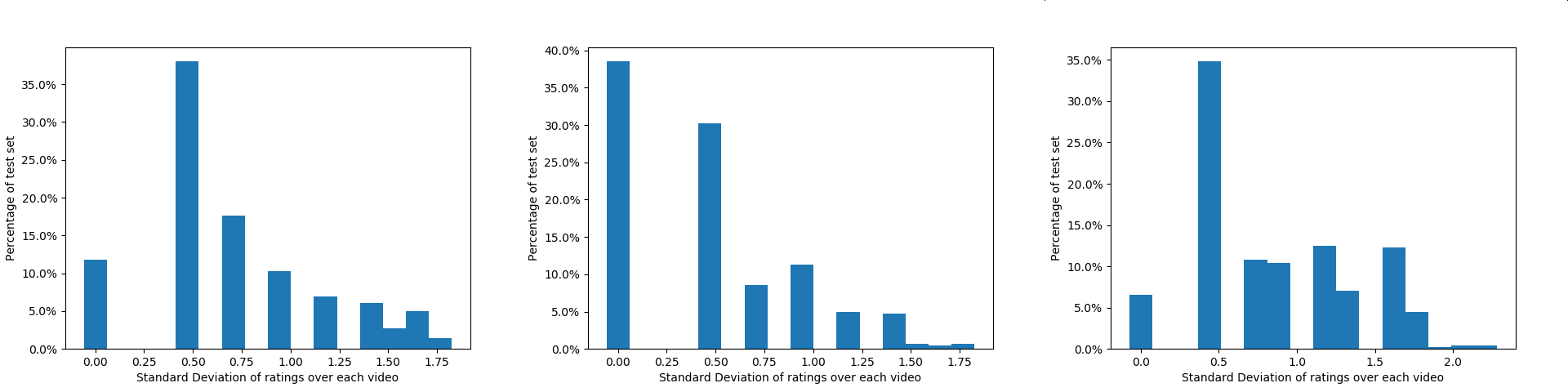}
            \caption{V2C-Completion task using our V2C-Transformer model.}
            \label{sfig:testb}
            \end{subfigure}\hfill
            \begin{subfigure}
            \centering
            \includegraphics[width=\linewidth]{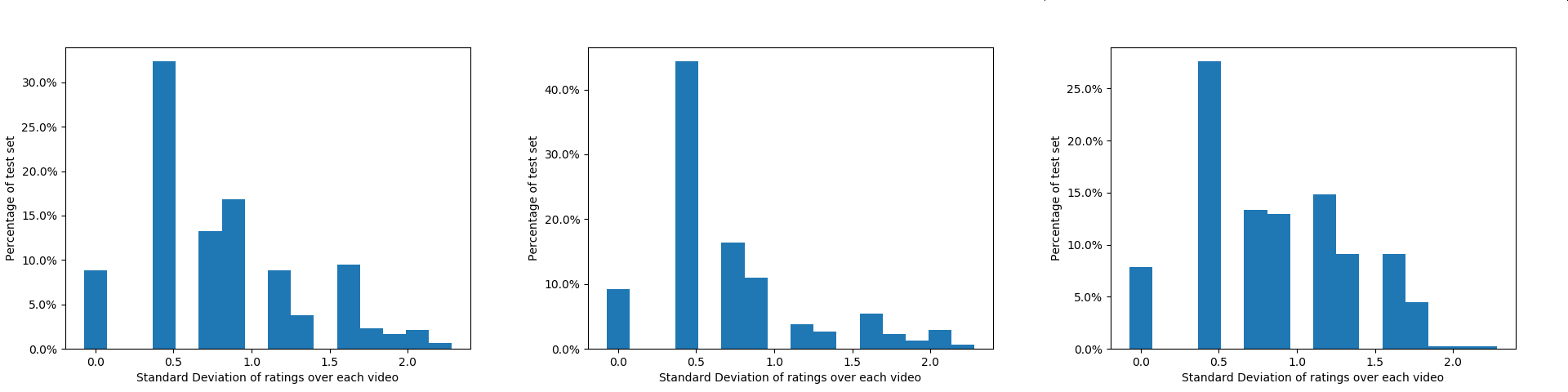}
            \caption{V2C-Completion task using the A\textsc{tt}E\textsc{nc}D\textsc{ec} model.}
            \label{sfig:testb}
            \end{subfigure}\hfill
            \begin{subfigure}
            \centering
            \includegraphics[width=\linewidth]{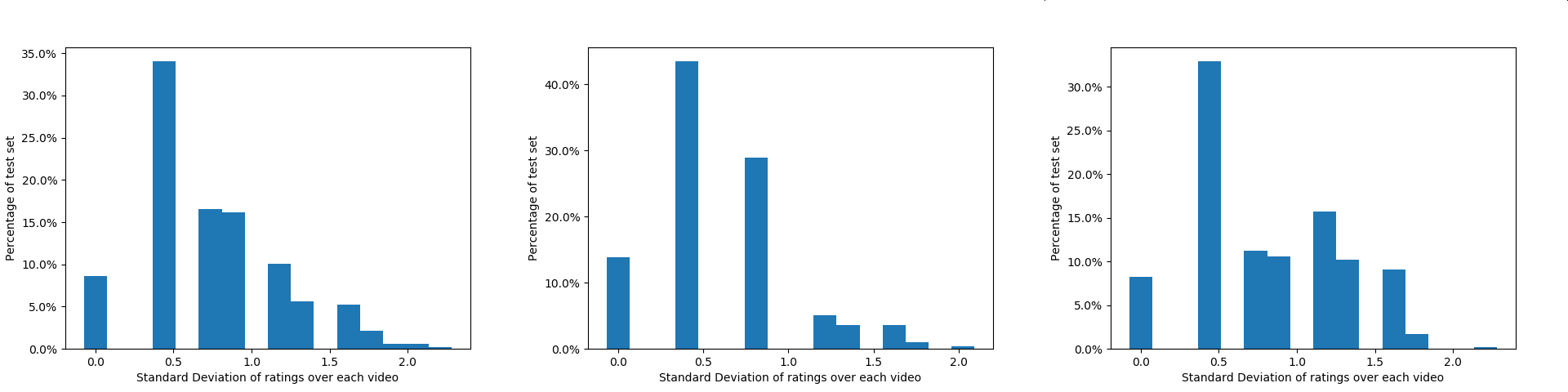}
            \caption{V2C-Generation task using our V2C-Transformer model.}
            \label{sfig:testb}
            \end{subfigure}\hfill
            \caption{Standard deviation histograms of human ratings across models and split 
            {\it(From left to right: Intention, Attribute, Effect)}. X-axis denotes standard deviation value and Y-axis denotes percentage of test set samples.}
            \label{fig:stdev}
        \end{figure*}

    \subsubsection{Inter-Rater Agreement Score}
    Inter-Rater Agreement Score is computed as the average of the percentage of raters for each sample that agree with the majority opinion.
    Let $m$ be the size of the test-set, and $n$ be the number of rating.
    Let $\mathcal{R}^{j}= \{r_1, \dots, r_n\}$ be the set of ratings for test sample $j$.
    Then the mode $r_{mode}$ is defined as the most frequently occurring (majority) rating in the set of ratings $\mathcal{R}^{j}$, i.e. $r_{mode} = \textsc{MODE}(\mathcal{R}^j)$.
    
    Inter-Rater Agreement Score $\mathbf{IRAS}_{agree}$ is the average percentage of raters that agree with the majority opinion $r_{mode}$:
    \begin{equation}
        \mathbf{IRAS} = 100 \times \sum_{j=1}^{m}\frac{\sum_{i=1}^{n}{\mathbb{I}(r_i = r_{mode})}}{n\times m}, 
    \end{equation}
    where $\mathbb{I}$ is the indicator function.
    
    \subsubsection{Smooth Inter-Rater Agreement Score}
    While IRAS acts as a good metric to find out how our dataset fares in terms of rater agreement, it suffers from a flaw.
    Irrespective of the value of ratings, the indicator function $\mathbb{I}$ returns 0 for the tuple of ratings ($1, 5$) as well as ($4, 5$), although the ratings of 4 and 5 are close to each other but 1 and 5 are opposite.
    So to avoid this, we replace the indicator function with a smooth exponential term.
    The smooth inter-rater agreement score is given by:
    \begin{equation}
        \mathbf{IRAS}_{smooth} = 100 \times \sum_{j=1}^{m}\frac{\sum_{i=1}^{n}{\big(\frac{1}{2}\big)^{|r_i - r_{mode}|}}}{n\times m}.
    \end{equation} 
    
    \subsubsection{Results}

    Table ~\ref{table:stdev} shows our analysis in terms of the three metrics described above. 
    Our V2C-Transformer architecture consistently outperforms the baseline model A\textsc{tt}E\textsc{nc}D\textsc{ec}~\cite{gao2017video} in all three metrics for each type of commonsense.
    This means that raters are more consistent with their ratings (in terms of deviation or agreement) for commonsense descriptions generated by our model.


%% file: tables/bert_ranking.tex
\begin{table}[t]
    \centering
    \small
    \begin{tabular}{c c c}
        \toprule
        \textbf{Commonsense Type} & \phantom{abc} & \textbf{Accuracy (\%)}  \\
        \toprule
        Intention   && 84.87  \\
        Effect      && 86.53  \\
        Attribute   && 87.23  \\
        \midrule
        Average     && 86.21  \\
        \bottomrule
    \end{tabular}
    \caption{Accuracy of our BERT model for next sentence prediction on the A\textsc{tomic} test dataset split}
    \label{tab:bert}
\end{table}

%% file: tables/v2cqa.tex
\begin{table*}[t]
    \centering
    \small
    \resizebox{\linewidth}{!}{
    \begin{tabular}{lclcl}
        \toprule
        \multicolumn{5}{c}{\raisebox{-\totalheight}{
        \includegraphics[width=15cm]{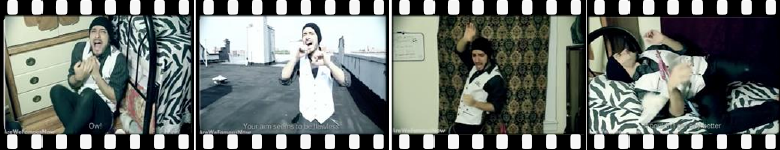}}} \\
        \midrule
        \textbf{Question Type} & \phantom{ab} & \textbf{Question} & \phantom{ab} & \textbf{Answer} \\
        \midrule
        Intention 
            && What might be the goal of the person? 
            && to record a music video \\
        Intention (Negative)
            && What could the person not want to achieve?
            && to bake a cake\\
        Intention (Action)
            && What prompts the person to do the action?
            && to express themselves\\
        Intention (Action, Negative)
            && What did not lead the person to act like that?
            && to feed the dog\\
         Intention (Why)
            && Why might the person be doing the action?
            && to entertain viewers\\
         Intention (Yes-No)
            && Does the person wish to express himself?
            && Yes \\
         Intention (Yes-No, Negative)
            && Does the person want to not get recognition?
            && No \\
        \midrule
        Effect
            && What will the person do after this?
            && puts the video on YouTube\\
        Effect (Negative)
            && What does not happen as a result? 
            && the person gets sad\\
        Effect (Action)
            && What does the dancing end up in?
            && becomes tired\\
        Effect (Action, Negative)
            && What will not happen due to the action? 
            && feels tense\\
        Effect (How)
            && How does the person feel after performing?
            && feels accomplised\\
        Effect (Yes-No)
            && Could the person put it on YouTube as a result?
            && Yes\\
        Effect (Yes-No, Negative)
            && Will the person not learn a new dance?
            && No\\
        \midrule 
        Attribute
            && What trait does the man possess?
            && musical\\
        Attribute (Negative)
            && What attribute does not match with the person?
            && angry\\
        Attribute (How)
            && How can the person be described?
            && entertaining\\
        Attribute (Action, How)
            && How can the dancing person be characterized?
            && rhythmic\\
        Attribute (Yes-No, Action)
            && Is the person who is singing smiling?
            && Yes\\
        Attribute (Yes-No)
            && Is the person entertaining?
            && Yes\\
        Attribute (Yes-No, Negative)
            && Is the person not tense?
            && Yes\\
        \bottomrule
    \end{tabular}
    }
    \caption{Examples of open-ended V2C-QA samples}
    \label{tab:v2cqa_data}
\end{table*}

%% file: tables/human_stats.tex
    \begin{table*}[!h]
    
    \begin{center}
    \resizebox{\linewidth}{!}{
    \begin{tabular}{cc l c c c c c c}
        &\hphantom& \textbf{Type} & \multicolumn{2}{c}{\textbf{Std. Dev (\%) $\downarrow$}} &  \multicolumn{2}{c}{\textbf{IRAS}(\%) $\uparrow$} & \multicolumn{2}{c}{\textbf{smooth-IRAS} (\%) $\uparrow$}   \\
        & & & A\textsc{tt}E\textsc{nc}D\textsc{ec} & V2C-Transformer & A\textsc{tt}E\textsc{nc}D\textsc{ec} & V2C-Transformer & A\textsc{tt}E\textsc{nc}D\textsc{ec} & V2C-Transformer \\
        \hline

        \multirow{3}{*}{\textbf{V2C-Completion}}  
        && {{\textbf{Intention}}} & 17.99 & 15.02 & 56.02 & 59.80 & 69.43 & 73.36  \\
        && {{\textbf{Effect}}} & 19.63 & 18.39  & 58.03 & 56.76 & 69.28 & 69.47\\
        && {{\textbf{Attribute}}} & 10.54 & 9.74  & 69.06 & 71.28 & 80.24 & 81.83\\
        \hline
        && {{\textbf{Average}}} & 16.05 & 14.38 & 61.04 & 62.61 & 72.98 & 74.89\\
        \hline
        
        \multirow{3}{*}{\textbf{V2C-Generation}} 
        && {{\textbf{Intention}}} & 17.60 & 16.27 & 57.84 & 58.47 & 70.66 & 72.10  \\
        && {{\textbf{Effect}}} & 18.54 & 17.56  & 56.69 & 57.40 & 69.54 & 70.21\\
        && {{\textbf{Attribute}}} & 15.42 & 13.16  & 59.80 & 62.25 & 73.51 & 76.12 \\
        \hline
        && {{\textbf{Average}}} & 17.19 & 15.66 & 58.11 & 59.37 & 71.24 & 72.81\\
    \end{tabular}
    }
    \end{center}
    \caption{A comparison of the statistics of human evaluation scores for both tasks using the baseline (A\textsc{tt}E\textsc{nc}D\textsc{ec} model vs. our model (V2C-Transformer)}
    \label{table:stdev}
\end{table*}